\documentclass[preprint,authoryear,nonatbib]{elsarticle}

\usepackage[T1]{fontenc}
\usepackage[utf8]{inputenc}
\usepackage{lmodern}
\usepackage{microtype}
\usepackage{amsmath,amssymb}
\usepackage{siunitx}
\usepackage{booktabs}
\usepackage{graphicx}
\usepackage{hyperref}
\hypersetup{hidelinks}
\usepackage{xcolor}
\usepackage{tabularx}
\usepackage{booktabs}
\usepackage{float}

\usepackage{comment}

\usepackage{geometry}
\usepackage{setspace}
\usepackage{url}

\usepackage{makecell} %
\usepackage{adjustbox}  

\graphicspath{{./}{./figures/}}

\usepackage{longtable,ltablex,booktabs,siunitx,makecell,array}

\hypersetup{
    colorlinks=true,
    linkcolor=blue,
    filecolor=blue,
    citecolor=blue,      
    urlcolor=blue,
}
\makeatletter
\let\c@author\relax
\makeatother

\usepackage[backend=biber,style=apa,doi=false,isbn=false,url=false]{biblatex}
\begin{document}

\begin{frontmatter}

\title{DAPFAM: A Domain-Aware Family-level Dataset to benchmark cross domain patent retrieval}

\author[1]{Iliass Ayaou\corref{cor1}%
\fnref{fn1}}
\ead{iliass.ayaou@insa-strasbourg.fr}
\author[1]{Denis Cavallucci}
\ead{denis.cavallucci@insa-strasbourg.fr}
\author[1]{Hicham chibane}
\ead{hicham.chibane@insa-strasbourg.fr}

\cortext[cor1]{Corresponding author}
\affiliation[1]{
organization={INSA Strasbourg, ICUBE Laboratory},
addressline={24 Bd de la Victoire},
postcode={67000},
city={Strasbourg},
country={France}
}

\begin{abstract}
Patent prior-art retrieval becomes especially challenging when relevant disclosures cross technological boundaries. Existing benchmarks lack explicit domain partitions, making it difficult to assess how retrieval systems cope with such shifts. We introduce \textsc{DAPFAM}, a family-level benchmark with explicit \textbf{IN}-domain and \textbf{OUT}-domain partitions defined by a new IPC3 overlap scheme. The dataset contains 1,247 query families and 45,336 target families aggregated at the family level to reduce international redundancy, with citation-based relevance judgments. We conduct 249 controlled experiments spanning lexical (BM25) and dense (transformer) backends, document- and passage-level retrieval, multiple query and document representations, aggregation strategies, and hybrid fusion via Reciprocal Rank Fusion (RRF). Results reveal a pronounced domain gap: \textbf{OUT-domain performance remains roughly five times lower} than IN-domain across all configurations. Passage-level retrieval consistently outperforms document-level, and dense methods provide modest gains over BM25, but none close the OUT-domain gap. Document-level RRF yields strong effectiveness–efficiency trade-offs with minimal overhead. By exposing the persistent challenge of cross-domain retrieval, \textsc{DAPFAM} provides a reproducible, compute-aware testbed for developing more robust patent IR systems. The dataset is publicly available on huggingface at \href{https://huggingface.co/datasets/datalyes/DAPFAM_patent}{this repository}.
\end{abstract}

\begin{keyword}
Information retrieval \sep Patent retrieval \sep Cross-domain retrieval \sep  Prior art search \sep BM25 \sep Passage retrieval \sep Reciprocal Rank Fusion
\end{keyword}

\end{frontmatter}

\newpage

\label{sec:introduction}
\section{Introduction}

Patent prior art retrieval is fundamental to innovation assessment, intellectual property strategy, and technological progress. Patent offices around the world, including the USPTO, EPO, WIPO, and JPO, collectively publish millions of technical disclosures that form a critical knowledge base for inventors, examiners, and researchers \parencite{fisher2001,Hallenborg2018}. However, the expanding volume and increasingly interdisciplinary nature of modern innovations create persistent challenges for effective prior art discovery, particularly when relevant patents span multiple technological domains.

\subsection{The Cross-Domain Challenge}
Contemporary innovation increasingly transcends traditional technological boundaries. A medical device may incorporate software algorithms, mechanical components, and telecommunications protocols; a pharmaceutical compound might rely on novel manufacturing processes from chemical engineering. However, existing patent retrieval benchmarks do not adequately address cross-domain scenarios in which query and target patents belong to different International Patent Classification (IPC) or Cooperative Patent Classification (CPC) codes \parencite{Risch2019}.

This limitation is problematic because cross-domain prior art search is both common and critical. Patent examiners routinely discover relevant prior art outside of an application's primary classification, and inventors must understand technological landscapes that extend beyond their immediate field to avoid infringement and build upon existing work \parencite{Abbas2014}. Traditional retrieval systems, optimized for in-domain matching, may miss these cross-domain connections due to vocabulary gaps, divergent technical terminologies, and varying document structures across technological domains.

\subsection{Dataset Limitations for Cross-Domain Evaluation}
Current patent retrieval datasets, including CLEF-IP \parencite{Piroi2011CLEFIP2R}, TREC Patent Track \parencite{Lupu2009}, MAREC \parencite{Marec2021}, and BigPatent \parencite{Sharma2019} advance the field, but share critical limitations for cross-domain research. Most focus on single jurisdictions, lack explicit domain partitioning for systematic evaluation, or omit family-level aggregation that reduces redundancy across international filings. Without standardized cross-domain evaluation protocols, researchers cannot rigorously compare retrieval methods under domain shift conditions or develop systems robust to technological diversity.

\subsection{Our Contribution}
We introduce \textsc{DAPFAM}, a family-level patent retrieval benchmark that addresses these gaps through systematic design choices. The dataset provides explicit \textbf{IN-domain} (shared IPC3 codes) and \textbf{OUT-domain} (no shared codes) evaluation partitions, enabling direct measurement of cross-domain retrieval difficulty. By aggregating patents at the family level and balancing queries across technological domains, \textsc{DAPFAM} supports reproducible research into domain-aware retrieval methods and hybrid fusion strategies.

\subsection{Novelty \& Significance}
This work makes three key contributions to patent retrieval research. 

\textbf{First}, we introduce the first family-level patent benchmark with explicit \emph{out-of-domain} evaluation partitions based on IPC3 overlap. Unlike existing datasets that treat all citations equally, \textsc{DAPFAM} systematically distinguishes between \textbf{IN-domain} (shared IPC3) and \textbf{OUT-domain} (no shared IPC3) retrieval scenarios, enabling direct measurement of cross-domain retrieval difficulty---a critical gap in current patent information retrieval evaluation.

\textbf{Second}, we provide comprehensive empirical analysis through systematic evaluation of key design choices: document versus passage granularity, query field combinations, passage aggregation strategies, and hybrid fusion. Rather than pursuing a multi-model comparison approach, we vary retrieval design decisions, providing insights into cross-domain retrieval behavior and best practices.

\textbf{Third}, we demonstrate that OUT-domain retrieval represents a fundamental challenge where dense methods lose their advantage over lexical approaches, with performance dropping approximately $5\times$ compared to IN-domain scenarios. This finding has important implications for patent search systems that must handle cross-domain prior art discovery.

\subsection{Problem Setting and Design Choices}
We evaluate at rank cutoff 100 with NDCG@100 and Recall@100 to balance ranking quality and coverage. The lexical backend anchors results with BM25~\parencite{Robertson2009}; the dense backend uses a single multilingual encoder chosen for its context length and moderate resource requirements. At passage level, documents are sliced into fixed-length windows and re-aggregated at the family level using four strategies: \textit{maxP}, \textit{avgP}, \textit{sumP}, and \textit{avg\_top3}. For hybrid retrieval, we adopt Reciprocal Rank Fusion (RRF)~\parencite{Cormack2009RRF}, which combines lexical and dense rankings. All sampling, indexing, and evaluation operate at the family level, thus reducing redundancy.

\subsection{Availability and Reproducibility}
We release the complete \textsc{DAPFAM} dataset at  \href{https://huggingface.co/datasets/datalyes/DAPFAM_patent}{this repository}, including: (1)~the family-level corpus with normalized text fields, (2)~balanced query sets with domain labels, (3)~citation-based relevance judgments with IN/OUT domain partitions, and (4)~comprehensive metadata enabling reproducible experiments and extended usage. The dataset supports multiple experimental configurations through varying query and corpus representations, passage slicing at multiple window sizes, and family-level aggregation strategies. All experimental configurations, hyperparameters, and computational details are fully documented.

\subsection{Structure of This Paper}
Section~\ref{sec:related} positions our contribution within patent benchmarks and information retrieval methods. Section~\ref{sec:data} details dataset construction, family-level unification, domain partitions, while section~\ref{sec:stats} offers a statistical overview of the dataset. Section~\ref{sec:methods} presents evaluation metrics, backends, passage aggregation, hybrid fusion, and the experimental design space. Section~\ref{sec:results} reports comprehensive results across granularities, query representations, passage effects, aggregation strategies, efficiency analysis, and cross-domain challenges. Section~\ref{sec:discussion} summarizes and discusses our key findings. Finally, Section~\ref{sec:conclusion} discusses future directions for expanding and refining this work to advance patent retrieval research.

\section{Background and Related Work}
\label{sec:related}

Patent retrieval sits at the intersection of computational linguistics, legal informatics, and classical information retrieval (IR). Its evolution has mirrored broader shifts in IR research, from keyword-based approaches to more sophisticated machine learning and neural methods, while simultaneously reflecting the unique demands imposed by legal and technical domains \parencite{Bonino2010}.

\paragraph{Early Milestones in Patent Retrieval}
Historically, patent searches were manual processes performed by examiners and specialized professionals. Before the widespread adoption of the internet, patent libraries often stored bulky volumes or microfilms, and searching involved flipping through classification indices or advanced notice bulletins \parencite{Bonino2010}. This labor-intensive process was feasible only for smaller sets of documents or narrower queries. As digital technology emerged, key patent offices such as the USPTO and EPO began providing limited online search interfaces. However, these initial online systems were often rudimentary, relying heavily on Boolean search operators and basic keyword matching. Such an approach frequently yielded either too many false positives (due to overly broad synonyms) or missed relevant documents if terminologies did not overlap exactly \parencite{Khode2017}.

The NTCIR Workshop Series was one of the earlier forums that attempted to systematize research on patent retrieval, particularly focusing on Japanese and English patents \parencite{Iwayama2004}. Each iteration of the workshop introduced specific tasks (like invalidity search, classification, and cross-lingual retrieval), encouraging participants to compare retrieval algorithms on shared datasets.

\subsection{Classical IR Approaches and the Emergence of Query Expansion}
Initial computational methods for patent retrieval often drew inspiration from standard text retrieval techniques, such as TF-IDF weighting \parencite{Robertson2004} and vector space models \parencite{Takaki2004}. In these methods, each patent document was treated like a standard text, albeit more technical in content. Such an approach worked reasonably well when searching within a single domain or language. But as the size and diversity of patent databases increased, it became apparent that domain-specific jargon and the multi-lingual aspects of patent filings posed additional hurdles \parencite{Bashir2010}.

\paragraph{Lexical Matching and BM25}

Among classical IR models, BM25 \parencite{Robertson2009} has been a workhorse in academic and industrial systems due to its solid balance between efficiency and effectiveness \parencite{Klampanos2009}. BM25 uses term frequency and inverse document frequency while capping the impact of extremely long documents. Patent text, however, can be extremely lengthy, tens or even hundreds of thousands of tokens, challenging the assumptions on which BM25's parameter tuning is based \parencite{GolestanFar2015}.

\paragraph{Query Expansion and Relevance Feedback}
To enhance recall for patent search, query expansion became a central strategy. Since the missing or mismatched terms between a query and a relevant document can be very large in the patent domain, expansions that capture domain-specific synonyms, morphological variants, or classification-based keywords significantly improve retrieval performance \parencite{GolestanFar2015}. Citation-based expansions further leverage the network of references among patents, injecting terms from closely linked prior art into the original query \parencite{Mahdabi2014,Lee2022}.

\paragraph{Incorporating Patent-Specific Metadata}
Patents differ from typical unstructured corpora because they come with structured metadata: classification codes (IPC, CPC), publication dates, priority dates, inventor names, applicant details, and forward/backward citations. Studies have shown that leveraging classification information (e.g., using IPC codes to guide query expansions) can yield more domain-relevant matches \parencite{Giachanou2015}. Moreover, citation-based retrieval and re-ranking schemes exploit the fact that patents cite earlier documents either as prior art references or as legal disclaimers \parencite{Krestel2021}. These citations are powerful signals of relevance, albeit an imperfect proxy, since not all relevant patents are cited. Nonetheless, they remain highly valued for tasks like prior art search, technological community detection  \parencite{Abbas2014} and patent value identification \parencite{LIU2023103327}.

\subsection{Neural Methods and Semantic Text Representations}
With the recent advancements in natural language processing (NLP), especially the emergence of deep learning architectures, new horizons have opened for patent IR. Traditional bag-of-words or n-gram methods are increasingly giving way to neural embeddings that capture semantic nuances. Techniques such as Word2Vec \parencite{word2vec2013} and Doc2Vec \parencite{doc2vec2014} paved the way for patent-specific embeddings, which can reduce the lexical gap by learning domain-centric vector representations \parencite{Risch2019}.

\paragraph{Transformer-Based Models}
Transformer architectures based models, including BERT \parencite{devlin2019} and domain-adapted variants, are particularly influential in capturing contextual semantics. Researchers have begun fine-tuning these models specifically on patent corpora \parencite{BEKAMIRI2024123536,STAMATIS2024102282}, leading to improved phrase-level matching, better handling of the hierarchical structure of claims, and more robust cross-domain retrieval capabilities. However, these models can be difficult to deploy for entire patent documents, since some patents exceed tens or hundreds of pages. 
\paragraph{Hybridization via reciprocal rank fusion (RRF).}
rank-based fusion is remarkably effective and robust across collections. We employ reciprocal rank fusion~\parencite{Cormack2009RRF}, \(F(d)=\sum_j \frac{1}{K+\mathrm{rank}_j(d)}\), selecting \(K\) per setting. Two operational regimes matter in practice: (i) \emph{passage-capable} fusion (both components built over passages) and (ii) \emph{doc-only} fusion (components over documents). The latter trades a small loss of headroom for markedly lower indexing cost and latency, which is attractive under tight budgets. Our experiments quantify both regimes under identical protocols.

\paragraph{Passage aggregation strategies.}
Passage-level retrieval requires consolidating evidence to the family (document) level.
Passage-level retrieval (splitting documents into sections or chunks) is thus an active area, with methods like ``maxP'' or ``firstP'' \parencite{Dai2019} scoring each chunk and aggregating the results to represent the patent as a whole.
We study three strategies that are frequently used---\emph{maxP} (highest score), \emph{avgP} (average of all passage scores), and \emph{sumP} (sum of all scores). We derive a variant from \emph{avgP} that we call \emph{avg\_top3} with $N{=}3$.

\paragraph{Domain Adaptation and Cross-Domain Challenges}
Even with the advanced language modeling capabilities of transformers, cross-domain retrieval in patent corpora remains a complex task. For instance, an invention described under the medical device classification might borrow concepts from software or mechanical engineering. A naive retrieval system trained on a single domain can struggle to recognize relevant prior art from neighboring fields. Research in domain adaptation explores ways to mitigate performance drops when query and target documents belong to distinctly different technical categories. Thus, explicit labeling of in-domain vs. out-of-domain relationships in a benchmark dataset is crucial for investigating these out-of-domain retrieval issues.

\subsection{Positioning DAPFAM vs Existing Patent Retrieval Datasets}

While existing datasets have advanced the field significantly, each has limitations that DAPFAM addresses through systematic design choices.

DAPFAM addresses critical gaps in existing patent retrieval benchmarks through three key design decisions. First, family-level aggregation reduces redundancy from multiple filings of the same invention across jurisdictions. Second, explicit domain partitioning based on IPC3 classification overlap enables systematic cross-domain evaluation, a capability absent from current datasets. Third, balanced query sampling across technological domains ensures comprehensive coverage while maintaining experimental tractability.

This systematic approach fills a methodological void in patent retrieval evaluation. While existing datasets provide valuable citation-based relevance judgments, none systematically distinguish between in-domain and cross-domain retrieval scenarios. This limitation has prevented rigorous analysis of domain transfer effects, despite their practical importance for patent search systems that must handle queries spanning multiple technological areas.

\textbf{CLEF-IP} provides large-scale EPO data with rigorous evaluation protocols, primarily targeting classification and retrieval tasks~\parencite{Piroi2011CLEFIP2R,Roda2010}. However, its European focus limits technical diversity, and it lacks explicit cross-domain evaluation capabilities despite containing IPC classification codes.

\textbf{TREC Patent Track} provided standardized prior art search evaluation with ~100 topics, but focused primarily on chemistry patents and lacked balanced domain coverage~\parencite{Lupu2009}.

\textbf{MAREC} offers massive scale (19M documents) and multi-jurisdictional coverage but requires extensive preprocessing and lacks standardized queries or domain labels~\parencite{Marec2021}.

\textbf{BigPatent} provides 1.3M USPTO documents for summarization tasks but lacks query sets or relevance judgments for retrieval evaluation~\parencite{Sharma2019}.

\textbf{WIPO Alpha} supports multi-lingual classification tasks but suffers from inconsistent document structures and lacks domain-aware retrieval capabilities~\parencite{Fall2003}.

\textbf{NTCIR} collections provided Japanese and USPTO patent retrieval tasks but suffered from narrow domain coverage and lack of cross-domain evaluation frameworks~\parencite{Takaki2004,Iwayama2004}.

These limitations motivate DAPFAM's systematic design: family-level aggregation reduces cross-jurisdictional redundancy while maintaining global coverage, explicit domain partitioning via IPC3 overlap scheme enables rigorous cross-domain evaluation and balanced query sampling ensures comprehensive technological representation.
    \paragraph{Family-Level Aggregation with Full Text:} Each query and target patent is aggregated at the family level, consolidating all textual fields (title, abstract, claims, description) to reduce duplicates and unify the invention's representation. Large corpora like MAREC do not come with pre-aggregated family-level texts or standardized queries. Having a ready to use dataset reduces the preprocessing overhead.
    \paragraph{Manageable Size for Passage-Level Research:} While sufficiently extensive for varied IR tasks, the dataset remains manageable for passage-splitting experiments. Datasets like MAREC are extremely large and unwieldy for smaller labs, whereas BigPatent lacks the pre-defined queries and domain labels needed for advanced retrieval tasks.

Below is a comparative table that encapsulates the key aspects of each existing patent retrieval dataset versus our proposed dataset. This table is designed to clearly illustrate the gaps in current resources and highlight the added value of our approach.

\begin{table}[!ht]
\centering
\footnotesize
\begin{tabularx}{\textwidth}{|X|X|X|X|X|X|}
\hline
\textbf{Dataset} & \textbf{Jurisdiction \& Scope} & \textbf{Query Set \& Domain Coverage} & \textbf{Domain Labeling \& Retrieval} & \textbf{Data Aggregation} & \textbf{Usability / Preprocessing Requirements} \\
\hline
CLEF-IP & Primarily EPO & No built-in query set; primarily for retrieval /classification & Lacks explicit in-/out-of-domain labeling & Document-level & Well-structured but limited by jurisdiction and dated coverage \\
\hline
TREC Patent Track & Primarily USPTO (with some EPO/WIPO); chemistry-focused & $\sim$1000 topics; narrow domain (primarily chemistry) & No explicit cross-domain labeling & Document-level & Standard IR metrics but limited query diversity and domain balance \\
\hline
MAREC & Global (EPO, WIPO, USPTO, etc.); massive scale & Raw corpus; no standardized query set; ad hoc domain labels & Lacks standardized domain labeling for cross-domain retrieval & Raw document-level; no family aggregation & Extensive preprocessing required; resource intensive for many labs \\
\hline
BigPatent & USPTO patents; broad coverage for summarization & Designed for summarization; lacks official query set and IR-specific labels & No domain-aware retrieval structure & Document-level with segmented fields & Ready for summarization studies but requires custom query design for IR experiments \\
\hline
WIPO Alpha & WIPO filings; multi-lingual with inconsistent formats & No standardized query set; partial focus on classification tasks & Lacks unified, domain-aware labeling & Document level with schema inconsistencies  & Extra preprocessing needed to standardize document structures \\
\hline
NTCIR Data & JPO and USPTO patents; narrow jurisdiction focus & Tailored for classification/invalidity searches and claim based IR & No explicit cross-domain partitioning & Document-level & Valuable for specific tasks but limited for broader, balanced retrieval studies \\
\hline
\textbf{DAPFAM (ours)} & \textbf{Global family-level patents in English; multi-jurisdiction scope} & \textbf{1,247 balanced queries covering diverse domains with robust citation networks} & \textbf{Explicit in-domain vs. out-of-domain labeling (via IPC codes) enabling cross-domain experiments} & \textbf{Family-level aggregation consolidating full text (title, abstract, claims, description)} & \textbf{Ready-to-use and manageable; minimal preprocessing required for both document- and passage-level IR tasks} \\
\hline
\end{tabularx}
\caption{Comparison of Existing Patent Retrieval Datasets and DAPFAM}
\label{tab:dataset_comparison}
\end{table}

By addressing these concerns, we aim to provide a resource that not only supports prior art search but also serves as a launchpad for more advanced retrieval studies, particularly for cross-domain queries. The next section details the step-by-step process by which we constructed this dataset, explaining how we selected queries, designated relevance, and labeled domain overlaps.

\section{Construction of DAPFAM}
\label{sec:data}
The design of our new patent retrieval dataset centers on a \textbf{three-step medallion architecture} (bronze, silver, gold) that ensures progressive filtering and curation of patent data from raw downloads to a final, user-ready format. This section outlines each major step, emphasizing data source selection, query creation, citation-based relevance judgment, and our in-domain/out-of-domain labeling strategy.

\begin{figure}[H]
\begin{center}
 \includegraphics[width=15cm,height=13cm]{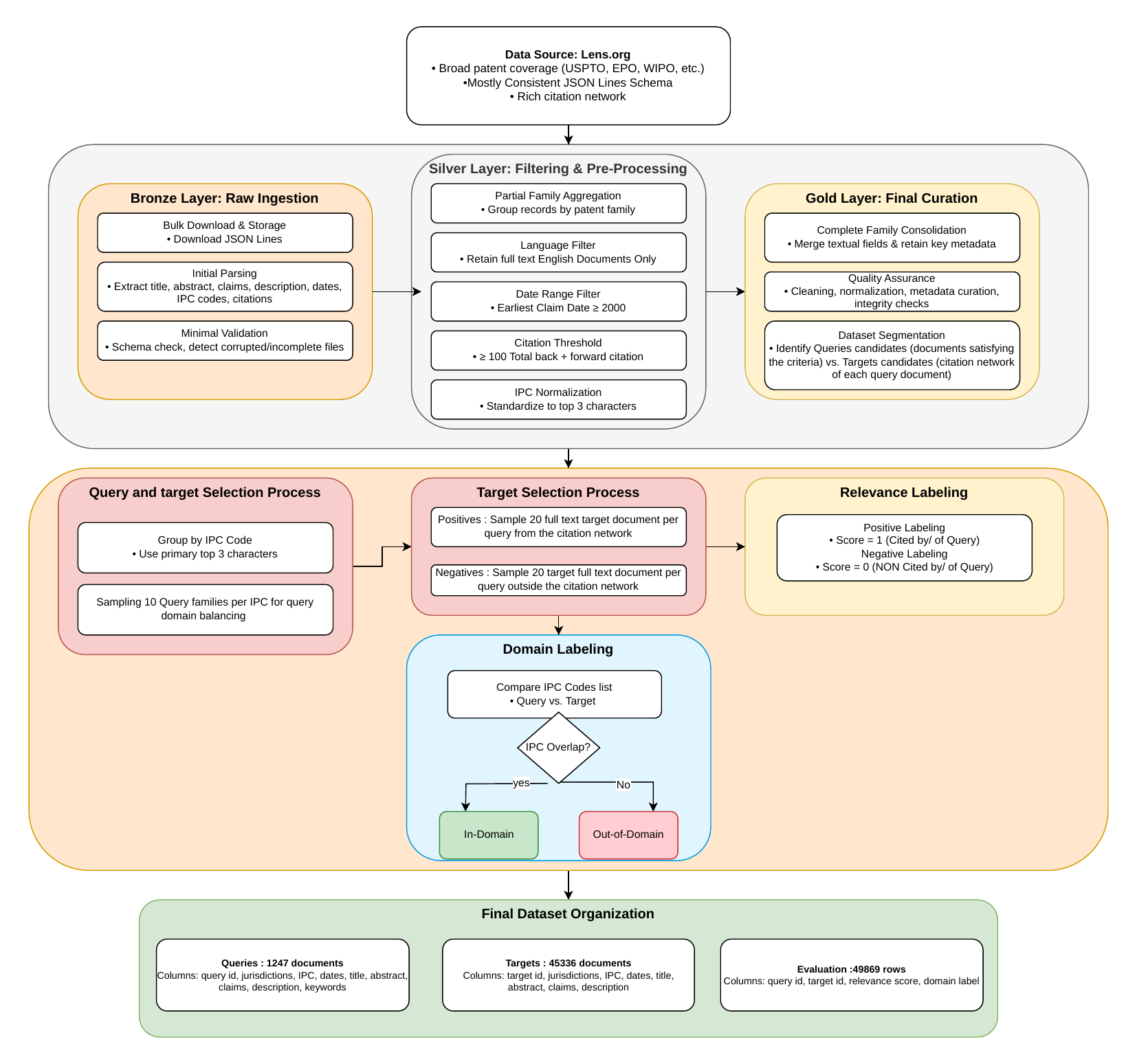}
\caption{Overview of the dataset construction steps.} \label{methodo_datset}
\end{center}
\end{figure}

\subsection{Data Source: Lens.org}

We selected \textbf{Lens.org} as our primary data provider based on four key advantages for patent retrieval research. \textbf{First}, comprehensive citation networks: the Lens aggregates forward and backward citations from multiple patent offices, providing rich citation graphs essential for relevance labeling. \textbf{Second}, global jurisdiction coverage: unlike single-office datasets, the Lens includes patents from USPTO, EPO, WIPO, and numerous national offices, enabling cross-jurisdictional family aggregation. \textbf{Third}, standardized data format: JSON lines output follows consistent schemas, reducing preprocessing complexity compared to heterogeneous XML formats. \textbf{Fourth}, research accessibility: transparent bulk download policies and academic pricing make large-scale corpus construction feasible for university research.

\subsection{Three-Step Medallion Architecture}

\paragraph{Bronze Layer (Raw Ingestion)}
The bronze layer establishes the foundation through systematic data acquisition and initial validation:
\begin{itemize}
    \item[] \textbf{Bulk Download and Storage:} We collected raw patent records in JSONL format from Lens.org. Each record represents an individual patent publication with complete bibliographic and textual content.
    \item[] \textbf{Initial Parsing:} Automated scripts extracted core textual fields (\texttt{title\_en}, \texttt{abstract\_en}, \texttt{claims\_text}, \texttt{description\_en}) and essential metadata (earliest claim date, jurisdiction, IPC classifications, citation lists).
    \item[] \textbf{Minimal Validation:} Schema conformity checks detected corrupted or incomplete records, ensuring all retained entries adhered to Lens.org's standardized data structure.
\end{itemize}

\paragraph{Silver Layer (Filtering and Preprocessing)}
The silver layer applies systematic filtering criteria to ensure data quality and experimental validity:
\begin{itemize}
    \item[] \textbf{Family-Level Grouping:} Patent publications belonging to the same simple family were aggregated while preserving individual member records. This intermediate step prepared comprehensive family-level data for final consolidation.
    \item[] \textbf{Language Standardization:} Only patent families with complete English text across all major fields (title, abstract, claims, description) were retained. Families with partial translations or missing English content were excluded to ensure uniform linguistic processing.
    \item[] \textbf{Temporal Filtering:} We restricted inclusion to patent families with earliest claim dates on or after January 1, 2000. This constraint ensures coverage of modern technological developments and sufficient temporal span for robust citation network formation.
    \item[] \textbf{Citation Network Threshold:} Patent families required a minimum of 100 combined forward and backward citations to qualify as potential queries. This threshold ensures meaningful citation-based relevance signals and excludes families with insufficient prior art connections.
    \item[] \textbf{IPC Code Normalization:} International Patent Classification codes were standardized to three-character prefixes (e.g., \texttt{A61} for medical/veterinary sciences, \texttt{H04} for telecommunications). This granularity provides the foundation for systematic domain partitioning.
\end{itemize}

\paragraph{Gold Layer (Final Dataset Curation)}
The gold layer produces the final, analysis-ready dataset through comprehensive consolidation and validation:
\begin{itemize}
    \item[] \textbf{Complete Family Consolidation:} Each patent family's textual content (title, abstract, claims, description) was merged into a single representative record using systematic aggregation rules: We take the earliest claim patent's id, Jurisdiction as a primary jurisdiction and date. array fields are aggregated while preserving distinct values only, and text fields retain the most recent version. Metadata consolidation preserved the earliest claim date and primary jurisdiction.
    \item[] \textbf{Quality Assurance:} Final data cleaning encompassed text normalization, metadata validation, and systematic integrity checks on a stratified sample of families to ensure consistency and completeness.
    \item[] \textbf{Query-Target Designation:} Curated families were partitioned into query candidates (meeting all filtering criteria) and target pools (drawn from citation networks), establishing the foundational structure for retrieval evaluation.
\end{itemize}

\subsection{Query and target Selection Process}

After obtaining fully curated family-level data from the gold layer, we grouped families by the first three characters of their \emph{primary} IPC code to ensure balanced domain representation. We targeted approximately 10 queries per IPC3 domain code for comprehensive technological coverage. In practice, domains with surplus qualifying candidates required random sampling, while others yielded fewer than 10 representatives. This balanced selection process produced \textbf{1,247 query families}, each constituting a distinct query entry in the final evaluation set.

For target families, we sampled approximately 20 full-text family documents per query from citation networks. This target count was selected to maintain dataset manageability for passage-level retrieval experiments while ensuring uniform evaluation conditions across queries. Since citation filtering operated on metadata independent of target language or text completeness, this sampling strategy provides consistent target pool sizes for robust comparative evaluation.

\subsection{Assigning Relevance via Citation Links}

Patent retrieval evaluation commonly employs examiner citations as relevance proxies, particularly for prior art search scenarios. We adopted this established citation-based labeling methodology for \textsc{DAPFAM}.

For each query family, we assigned binary relevance scores: any family appearing in the query's forward or backward citation network received a relevance score of 1, reflecting examiner-validated technical relationships. This labeling scheme aligns with patent examination practices where cited prior art represents legally and technically relevant precedents for novelty assessment.

To support supervised learning and enable negative sampling for ranking model training, we randomly sampled approximately 20 non-cited families per query as negative examples (relevance score = 0).

Table~\ref{tab:relevance_labeling} shows a simplified illustration of how citation-based labeling is assigned for a hypothetical query family (Q123).

\begin{table}[!ht]
\centering
\caption{Illustration of Citation-Based Relevance Assignment}
\label{tab:relevance_labeling}
\begin{tabular}{llll}
\toprule
\textbf{Query} & \textbf{Target} & \textbf{Citation Relationship} & \textbf{Relevance Score} \\
\midrule
Q123           & T456            & Q123 cites T456                & 1 \\
Q123           & T789            & T789 cites Q123                & 1 \\
Q123           & T999            & No citation link               & 0 \\
\bottomrule
\end{tabular}
\end{table}

\subsection{Domain Labeling: In-Domain vs. Out-of-Domain}

A core innovation of \textsc{DAPFAM} is systematic \textbf{domain partitioning} that enables direct evaluation of cross-domain retrieval difficulty. For each query-target pair, we classify the relationship as \textbf{IN-domain} or \textbf{OUT-domain} based on IPC3 classification overlap.

\paragraph{IPC3 Overlap Definition.} We extract the first three characters of all IPC codes assigned to each patent family (e.g., \texttt{A61} from \texttt{A61B17/00}). A target is labeled \textbf{IN-domain} if it shares at least one IPC3 code with the query family; otherwise, it is \textbf{OUT-domain}.

\begin{itemize}
    \item[] \textbf{IN-Domain Example:} Query IPC3 = \{\texttt{A61}, \texttt{H04}\}, Target IPC3 = \{\texttt{A61}, \texttt{B29}\} $\implies$ Overlap = \{\texttt{A61}\} $\implies$ IN-domain.
    \item[] \textbf{OUT-Domain Example:} Query IPC3 = \{\texttt{H04}\}, Target IPC3 = \{\texttt{G06}\} $\implies$ No overlap $\implies$ OUT-domain.
\end{itemize}

\paragraph{Rationale for IPC3 Granularity.} The three-character level balances domain specificity with evaluation coverage. Higher levels (e.g., \texttt{A} = ``Human Necessities'') are excessively broad, while lower levels (e.g., \texttt{A61B} = ``Diagnosis; Surgery; Identification'') create overly narrow partitions that reduce statistical power. IPC3 codes represent coherent technological domains (e.g., \texttt{A61} = medical/veterinary, \texttt{H04} = telecommunications, \texttt{G06} = computing) while maintaining sufficient query-target pairs in each partition for robust evaluation.

\paragraph{Multi-Code Inclusion.} We include both primary and secondary IPC classifications to ensure OUT-domain targets are genuinely distant from query domains, creating a challenging cross-domain retrieval scenario that reflects real-world prior art search complexity.

\subsection{Final Dataset Organization}

The \textsc{DAPFAM} dataset is organized into three complementary files supporting flexible experimental design: \textbf{Queries} (1,247 balanced patent families), \textbf{Targets} (45,336 candidate families), and \textbf{Evaluation} (query-target pairs with relevance and domain labels). This modular structure enables researchers to experiment with different query representations, corpus subsets, and evaluation metrics while maintaining consistent ground truth annotations.

\begin{table}[!ht]
\centering
\caption{Columns in \texttt{Queries / Targets}}
\label{tab:queries_targets_csv}
\begin{tabularx}{\textwidth}{>{\ttfamily}l X}
\toprule
\textbf{Column} & \textbf{Description} \\
\midrule
query\_id / relevant\_id  & Unique identifier for each query/target family. \\
earliest\_claim\_jurisdiction & Jurisdiction of the patent with the earliest claim date in the family. \\
jurisdiction & Complete list of jurisdictions covered by this patent family. \\
ipcr\_codes\_str & Raw string containing full IPC classifications assigned to the family. \\
earliest\_claim\_date & Date of the earliest claim in the family (chronological reference point). \\
earliest\_claim\_year & Numeric year extracted from earliest claim date. \\
classifications\_ipcr\_list\_first\_three\_chars\_list & IPC3 codes (first three characters) used for domain classification. \\
title\_en & Patent family title in English. \\
abstract\_en & Patent family abstract in English. \\
claims\_text & Complete claims text from the patent family. \\
description\_en & Patent family description text in English. \\
abstract\_keywords (queries only) & Keywords automatically extracted from abstract text. \\
\bottomrule
\end{tabularx}
\end{table}

\begin{table}[!ht]
\centering
\caption{Columns in \texttt{Evaluation}}
\label{tab:eval_csv}
\begin{tabularx}{\textwidth}{>{\ttfamily}l X}
\toprule
\textbf{Column} & \textbf{Description} \\
\midrule
query\_id & Identifier linking to \texttt{Queries}. \\
relevant\_id & Identifier linking to \texttt{Targets}. \\
relevance\_score & Binary or numeric scale indicating relevance (1 for cited prior art, 0 for randomly sampled negatives). \\
domain\_rel & Label for in-domain or out-of-domain, based on IPC overlap at the first-three-character level. \\
\bottomrule
\end{tabularx}
\end{table}

After applying the above filters and sampling strategies, we arrived at:
\begin{itemize}
    \item[] \textbf{1,247} query families (each with a thorough set of metadata and full English text).
    \item[] \textbf{45,336} target families, which constitute cited patents (relevant) plus random negatives.
    \item[] For each query, about \textbf{20 relevant} (positive) documents were sampled from all cited families, and \textbf{20 negative} (non-cited) documents were added, in order to ease the use of the dataset for model training using contrastive learning \parencite{zhang2022contrastive}.
\end{itemize}

By presenting a balanced approach, explicit domain labeling, moderate yet substantial data size, multi-jurisdiction coverage, and family-level aggregation, this dataset sets the stage for a broad spectrum of IR experiments. The next section describes a variety of statistical analyses that reveal further insights into domain distribution, textual complexity, citation patterns, and jurisdiction coverage.

\section{Dataset Statistics and Analysis}
\label{sec:stats}

Before embarking on retrieval experiments, a thorough statistical overview helps to understand the dataset's composition and potential challenges. In this section, we examine domain distribution (via IPC codes), time ranges, citation patterns, text lengths, jurisdictions, and the in-domain vs. out-of-domain relevance split.

The dataset comprises \textbf{1,247 queries} and \textbf{45,336 target families}, connected by \textbf{49,869} evaluation records (approximately 20 positives + 20 negatives per query).

\subsection{Domain Distribution and Coverage}

\begin{figure}[!ht]
\begin{center}
 \includegraphics[width=15cm,height=11cm]{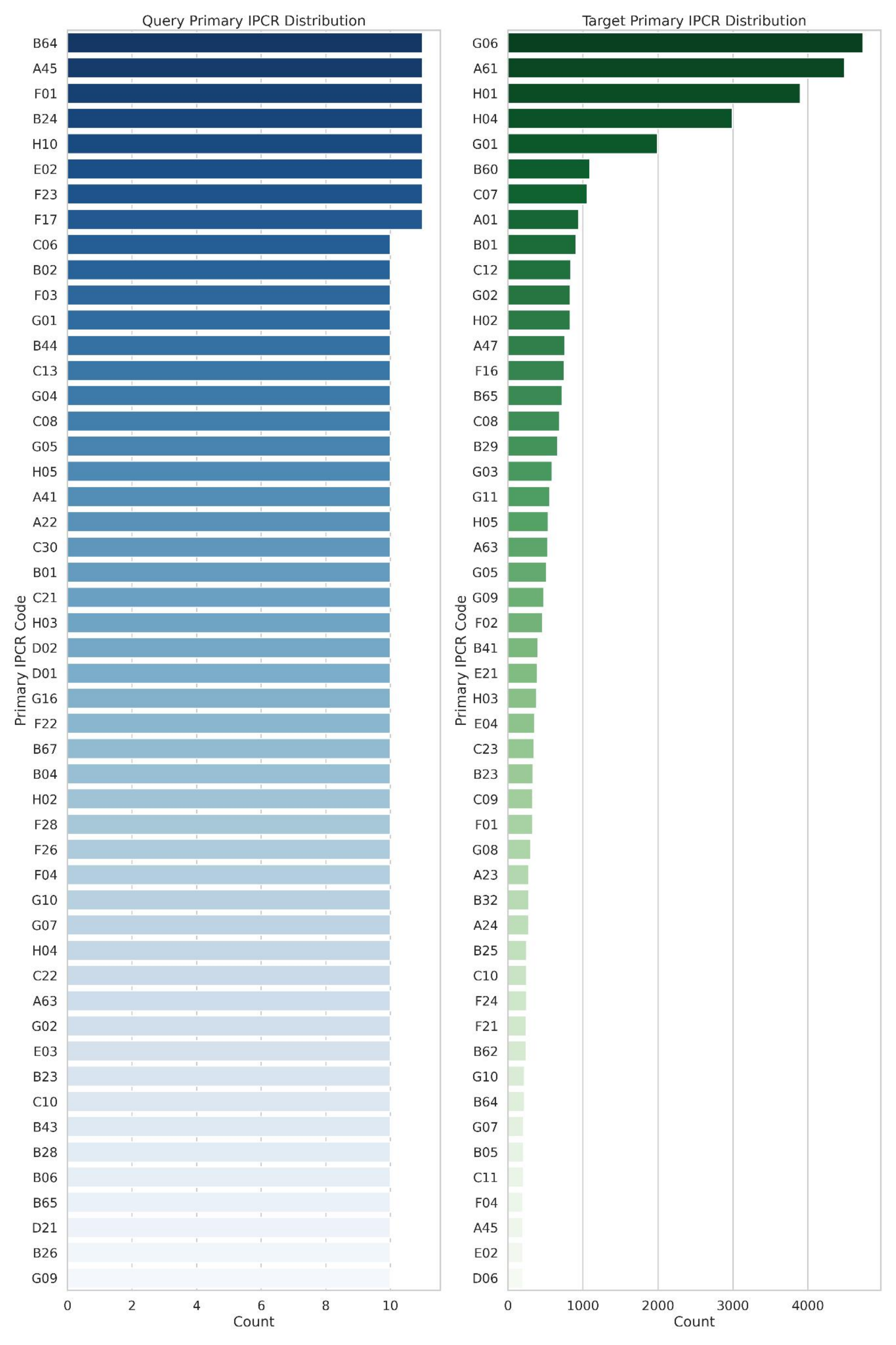}
\caption{Frequency of top 50 level 3 IPC domains for query and target documents.} \label{ipc_distrib}
\end{center}
\end{figure}
The dataset achieves broad technological coverage with representation from all major IPC3 sections. Figure~\ref{ipc_distrib} provides a detailed view of how queries and target documents are distributed among top 50 selected IPC classes. This diversity ensures that cross-domain retrieval experiments reflect realistic patent search scenarios spanning multiple technological boundaries.

\subsection{Earliest Claim Date Distribution}

\begin{figure}[!ht]
\begin{center}
 \includegraphics[width=15cm,height=15cm,keepaspectratio]{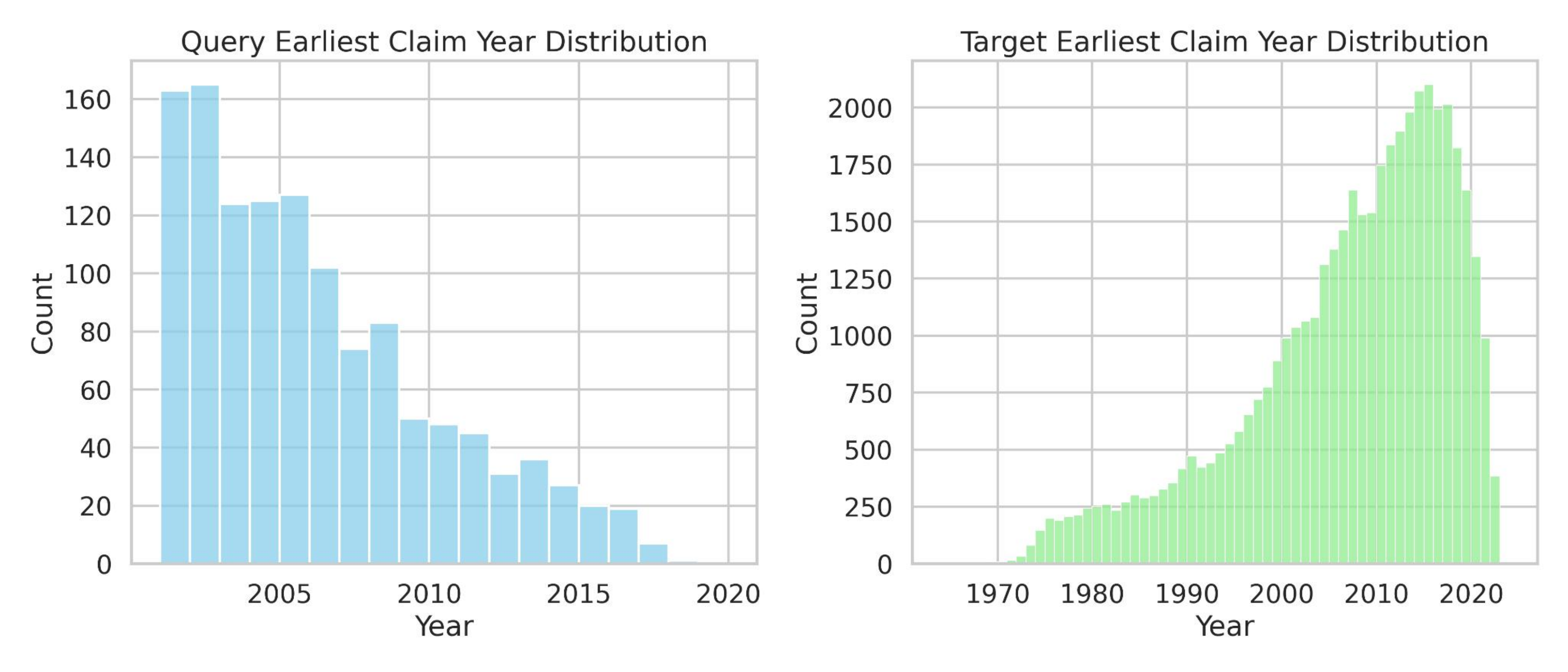}
\caption{Histograms for earliest claim years of queries vs. targets.} \label{query_target_distrib}
\end{center}
\end{figure}

All queries were filed on or after 2000, but their cited/cited by references (targets) cover a larger time frame. Figure~\ref{query_target_distrib} outlines the distribution of earliest claim years for queries and targets. While queries cluster around the 2000--2015 period, the target families display a wider spread.

The presence of older cited documents provides an opportunity to test retrieval strategies across varying publication eras, analyzing how changes in language usage or classification schema over time might affect IR performance.

\subsection{Temporal Coverage and Citation Patterns}

\begin{figure}[H]
  \centering
  \includegraphics[width=12cm,height=10cm]{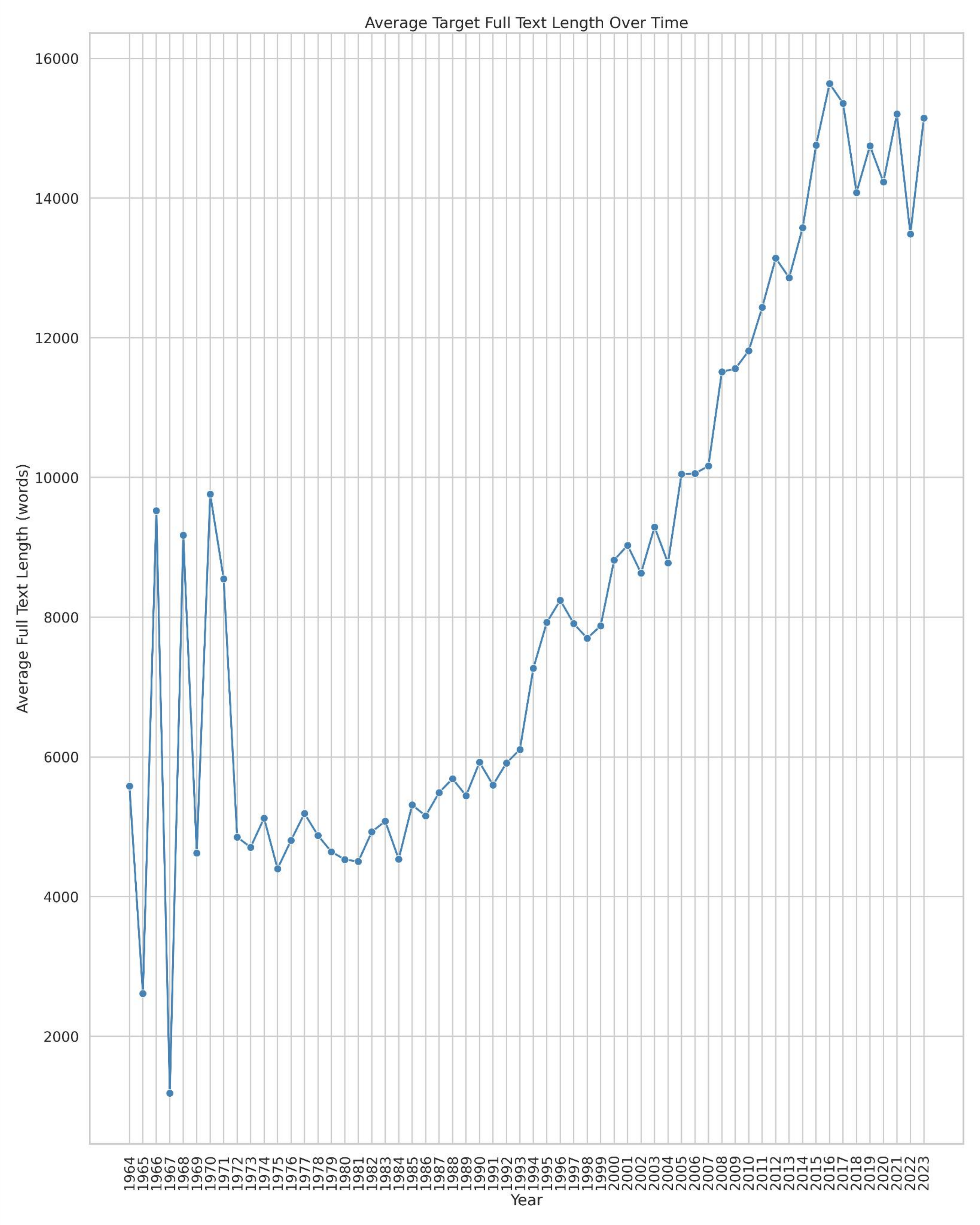}
  \caption{Average patent text length growth over time (1990–2020). Modern patents contain substantially more text, motivating passage-level retrieval approaches that can handle increased document complexity.}
  \label{fig:text-length}
\end{figure}

The temporal scope spans 1964–2023, with earliest claim dates for query families concentrated in the 2000–2015 period to ensure substantial citation networks for targets. Figure~\ref{fig:text-length} demonstrates the growth in patent text length over time, reflecting increased technical complexity and detailed disclosure requirements in modern patent applications.

\subsection{Text Length Analysis}

Patent documents can vary significantly in length, from concise short patents to voluminous multi-section specifications. Table~\ref{tab:query_target_len_stats} showcases descriptive statistics for the text length (in tokens) after merging each family's title, abstract, claims, and description.

\begin{table}[H]
\centering
\caption{Full Text Length Statistics for query and target documents}
\label{tab:query_target_len_stats}
\begin{tabularx}{\textwidth}{l *{2}{>{\centering\arraybackslash}X}}
\toprule
Statistic & \makecell{Query Full Text} & \makecell{Target Full Text} \\
\midrule
Count               & 1247   & 45336   \\
Mean                & 20448.05 & 11090.01 \\
Standard Deviation  & 30667.00 & 13768.34 \\
25th Percentile     & 7486   & 4544    \\
50th Percentile     & 12330  & 7432    \\
75th Percentile     & 21238.50 & 12552.50 \\
\bottomrule
\end{tabularx}
\end{table}

Queries, on average, are longer (mean \(\approx 20,448\) tokens) than targets (mean \(\approx 11,090\) tokens). This difference might be attributed to the fact that query documents are specifically chosen from highly cited families, which often contain more extensive specifications, multiple claim sets, and broader disclosures. Such length variation demands adaptive retrieval strategies, as naive keyword matches might be overwhelmed by the sheer volume of text.

All queries were filed on or after 2000, with citation networks extending to earlier decades, creating opportunities to analyze retrieval performance across varying publication eras and technological evolution.

\subsection{Jurisdiction Coverage}

\begin{table}[!ht]
\centering
\caption{Distribution of Query and Target Patents by Jurisdiction}
\label{tab:jurisdiction_coverage}
\begin{tabularx}{\textwidth}{l *{4}{>{\centering\arraybackslash}X}}
\toprule
Jurisdiction & \makecell{Number of\\Queries} & \makecell{Percentage\\(Queries)} & \makecell{Number of\\Targets} & \makecell{Percentage\\(Targets)} \\
\midrule
US     & 969   & 77.71\% & 27656 & 61.00\% \\
JP     & 111   & 8.90\%  & 5907  & 13.03\% \\
KR     & 30    & 2.41\%  & 1573  & 3.47\%  \\
DE     & 24    & 1.92\%  & 1880  & 4.15\%  \\
EP     & 23    & 1.84\%  & 1920  & 4.24\%  \\
GB     & 21    & 1.68\%  & 1143  & 2.52\%  \\
CN     & 15    & 1.20\%  & 1268  & 2.80\%  \\
FR     & --    & --      & 669   & 1.48\%  \\
SE     & 9     & 0.72\%  & --    & --      \\
AU     & 7     & 0.56\%  & --    & --      \\
NL     & 6     & 0.48\%  & --    & --      \\
TW     & --    & --      & 448   & 0.99\%  \\
IT     & --    & --      & 447   & 0.99\%  \\
Others & 32    & 2.57\%  & 2425  & 5.35\%  \\
\bottomrule
\end{tabularx}
\end{table}

Our approach to presenting only English text does not imply that all families originated in English-speaking countries. Rather, we unify the text at the family level, pulling English versions of patents that were also possibly filed in non-English jurisdictions at first. Table~\ref{tab:jurisdiction_coverage} highlights the top jurisdictions for the earliest claim date among both query and target families.

As expected, US documents dominate, reflecting global patent-filing behaviors where the USPTO is often prioritized and the inherent bias that spans from filtering on English text only. Japan (JP), Europe (EP), Korea (KR), Germany (DE), and China (CN) also appear prominently. This multi-national presence enables cross-jurisdiction analyses, even if the final text is in English.

\subsection{In-Domain vs. Out-of-Domain Relevance}

\begin{table}[ht]
\centering
\caption{Average Count and Percentage by Domain Relation}
\label{tab:domain_relevance_stats}
\begin{tabularx}{0.6\textwidth}{l c c}
\toprule
Domain Relation & Average Count & Percentage \\
\midrule
in\_domain  & 16.2169 & 73.92\% \\
out\_domain & 5.7381  & 26.08\% \\
\bottomrule
\end{tabularx}
\end{table}

A central innovation of this dataset is the labeling of each relevant target as \textbf{in-domain} or \textbf{out-of-domain}, based on the overlap of IPC3 codes. Table~\ref{tab:domain_relevance_stats} offers a snapshot of the counts:

We see that \textbf{about 26\%} of relevant documents do not share any (first-three-character) IPC code with the query. This subset is particularly valuable for studying cross-domain retrieval. Some queries exhibit higher rates of out-of-domain citations, especially in interdisciplinary areas or fields that rely on fundamental technologies from other domains.

\section{Evaluation \& Methods}
\label{sec:methods}

\subsection{Task Setup and Views}
Our retrieval unit is the \emph{patent family}. Given a query family $q$, the system ranks candidate families $d \in \mathcal{D}$; all sampling, indexing, and scoring operate at the family level to minimize redundancy across jurisdictions and near-duplicate publications. We evaluate on three disjoint subsets: \textsc{ALL} (complete test set), \textsc{IN} (candidates sharing at least one IPC3 code with $q$), and \textsc{OUT} (candidates with no shared IPC3 codes). 

We consider two retrieval granularities. In the \textbf{document} setting, each family is represented by a single consolidated text unit per field combination. In the \textbf{passage} setting, family text is segmented into fixed-size windows; passages receive independent scores that are subsequently aggregated to produce a single family-level score (Section~\ref{subsec:agg}). We systematically vary \textit{query\_view} and \textit{corpus\_view} to reflect practical field combinations (e.g., \textsc{Title+Abstract}, \textsc{Title+Abstract+Claims}, \textsc{Full Text}). Passage segmentation employs token windows of length $p \in \{64, 128, 256, 512, 1024, 2048, 4096, 8192\}$ with consistent stride and filtering parameters across backends. These \emph{derived} views (text representations and passage slices) are generated programmatically during experiments; only the base corpus and family-level relevance pairs are distributed.

\subsection{Metrics}\label{subsec:metrics}
We report \emph{Normalized Discounted Cumulative Gain} at rank 100 (NDCG@100) and \emph{Recall} at rank 100 (Recall@100), macro-averaged over queries within each evaluation subset (\textsc{ALL}/\textsc{IN}/\textsc{OUT}). Let $\{d_i\}_{i=1}^{100}$ denote the top-ranked results for query $q$ and $\mathrm{rel}(d) \in \{0,1\}$ represent the binary relevance label:
\begin{equation}
\mathrm{NDCG@100}(q) \;=\; \frac{1}{\mathrm{IDCG}_{100}(q)} \sum_{i=1}^{100} \frac{2^{\mathrm{rel}(d_i)}-1}{\log_2(i+1)}\,,
\end{equation}
\begin{equation}
\mathrm{Recall@100}(q) \;=\; \frac{\big|\{d \in \mathrm{Top}@100 : \mathrm{rel}(d)=1\}\big|}{\big|\{d:\mathrm{rel}(d)=1\}\big|}\,.
\end{equation}
We emphasize rank cutoff 100 to balance early-precision sensitivity with coverage requirements, reflecting real-world prior art workflows where examiners routinely examine results beyond the first page of rankings.

\subsection{Backends and Compute Envelope}\label{subsec:backends}
\textbf{Lexical Backend.} We employ a modern BM25 implementation (\texttt{bm25s}) using library default parameters ($k_1{=}1.2$, $b{=}0.75$) unless otherwise specified. BM25 has emerged as a robust and transparent lexical baseline~\parencite{Robertson2009}.  
\textbf{Dense Backend.} We utilize a single multilingual encoder (\texttt{Snowflake/snowflake-arctic-embed-m-v2.0}) across all dense retrieval experiments. Vector representations are quantized to \texttt{int8} precision to accommodate our computational constraints (24-core CPU, RTX 4090, 60GB RAM) while maximizing experimental breadth. This single-encoder approach eliminates model selection confounds and emphasizes systematic design decisions (granularity, field views, aggregation strategies). 

\subsection{Passage Slicing and Aggregation}\label{subsec:agg}
For a candidate family $d$ with passage set $P(d)$ and per-passage scores $\{s_p\}_{p \in P(d)}$, we evaluate four aggregation strategies to produce family-level scores:
\begin{align}
\textit{maxP:}\quad  s(d) &= \max_{p \in P(d)} s_p \label{eq:maxp}\\
\textit{avg\_top3:}\quad s(d) &= \frac{1}{3}\sum_{p \in \mathrm{Top3}(P(d))} s_p \label{eq:avgtop3}\\
\textit{avgP:}\quad s(d) &= \frac{1}{|P(d)|}\sum_{p \in P(d)} s_p \label{eq:avgp}\\
\textit{sumP:}\quad  s(d) &= \sum_{p \in P(d)} s_p \label{eq:sump}
\end{align}
where each strategy addresses different passage score distributions:
\begin{itemize}
\item \textbf{maxP}: Selects the highest-scoring passage, emphasizing peak relevance signals while avoiding dilution from irrelevant content.
\item \textbf{avg\_top3}: Averages the three highest-scoring passages, balancing signal strength with robustness to outliers.
\item \textbf{avgP}: Computes the mean across all passages, providing a global family-level relevance estimate.
\item \textbf{sumP}: Sums all passage scores, favoring longer documents with multiple relevant sections.
\end{itemize}
We fix $N=3$ for \textit{avg\_top3} to balance signal quality and statistical stability across patent families of varying lengths.

\subsection{Hybridization via Reciprocal Rank Fusion}\label{subsec:rrf}
To combine complementary retrieval signals, we employ \emph{Reciprocal Rank Fusion} (RRF)~\parencite{Cormack2009RRF}. Given component rankers indexed by $j$ and candidate family $d$, the fused score is:
\begin{equation}
F(d) = \sum_{j} \frac{1}{K + \mathrm{rank}_j(d)}, \label{eq:rrf}
\end{equation}
where $\mathrm{rank}_j(d)$ denotes the rank of family $d$ under component system $j$, and $K > 0$ is a smoothing constant.

We analyze two operational regimes: (i)~\textbf{passage-level hybrids} combining passage-level BM25 and dense components, and (ii)~\textbf{document-only hybrids} combining document-level components. The RRF constant $K$ is selected via grid search over $K \in \{10, 30, 60, 100\}$, optimizing NDCG@100 on the \textsc{ALL} subset. This approach yields $K = 30$ for passage-level hybrids and $K = 60$ for document-only hybrids.

\subsection{Design Space and Breadth}\label{subsec:design}
Our evaluation systematically explores the Cartesian product of the following design factors:
\begin{itemize}
  \item \textbf{Backend} $\in \{\textsc{BM25}, \textsc{Dense}\}$ (lexical vs. neural retrieval),
  \item \textbf{Granularity} $\in \{\textsc{Document}, \textsc{Passage}\}$ (indexing unit),
  \item \textbf{Query representation} $\in \{\textsc{Title}, \textsc{Title+Abstract}, \textsc{Title+Abstract+Claims}, \textsc{Keywords}\}$,
  \item \textbf{Corpus representation} $\in \{\textsc{Full Text}, \textsc{Title+Abstract}, \textsc{Title+Abstract+Claims}\}$,
  \item \textbf{Aggregation strategy} $\in \{\textit{maxP}, \textit{avg\_top3}, \textit{avgP}, \textit{sumP}\}$ (passage-level only),
  \item \textbf{Passage length} $p \in \{64, 128, 256, 512, 1024, 2048, 4096, 8192\}$ tokens (passage-level only).
\end{itemize}
This systematic exploration yields \textbf{249} unique experimental configurations (deduplicated by factor combinations). This breadth-focused approach, prioritizing design space coverage over model variety, enables principled, compute-constrained comparisons.

\subsection{Protocol and Implementation}\label{subsec:protocol}

\textbf{Indexing Protocol.} Document-level runs create one index entry per family per corpus view representation. Passage-level runs segment family text into windows while preserving family identifiers for subsequent aggregation.

\textbf{Scoring Implementation.} BM25 employs library default parameters unless explicitly noted. Dense embeddings are computed once using fixed random seeds and reused across all experimental runs to ensure consistency.

\textbf{Evaluation Procedure.} All systems are evaluated at rank cutoff $k = 100$, computing NDCG@100 and Recall@100 for each evaluation subset (\textsc{ALL}/\textsc{IN}/\textsc{OUT}). Execution timing and all configuration parameters are logged for provenance tracking.

\textbf{Efficiency Measurement.} Runtime analysis measures scoring computation and aggregation time only, excluding indexing and embedding preprocessing to focus on query-time performance trade-offs.

\begin{table}[!ht]
\centering\small
\caption{Notations table.}
\label{tab:notation}
\begin{tabular}{l l}
\toprule
\textbf{Symbol} & \textbf{Meaning} \\
\midrule
$q$ & query family \\
$d$ & candidate family; $P(d)$ its passages \\
$s_p$ & score of passage $p$; $s(d)$ aggregated family score \\
$\mathrm{rank}_j(d)$ & rank of $d$ under component $j$ \\
$K$ & RRF constant selected by grid search on \textsc{ALL} \\
$\mathrm{NDCG@100}$ & normalized discounted cumulative gain at 100 \\
$\mathrm{Recall@100}$ & recall at 100 \\
\textsc{ALL}/\textsc{IN}/\textsc{OUT} & evaluation subsets (domain partitions) \\
\textsc{document}/\textsc{passage} & retrieval granularity \\
$p$ & passage length in $\{64,\dots,8192\}$ \\
\bottomrule
\end{tabular}
\end{table}

\section{Results}
\label{sec:results}

This section presents comprehensive experimental results across 249 unique configurations, analyzing the systematic effects of backend choice, granularity, query representations, aggregation strategies, and hybrid fusion on retrieval effectiveness. Results are organized to examine: (1) single-backend performance comparisons, (2) query representation effectiveness, (3) passage length optimization, (4) aggregation strategy analysis, (5) hybrid fusion performance, and (6) efficiency-effectiveness trade-offs. All results report NDCG@100 and Recall@100 averaged across queries within evaluation subsets.
\subsection{Backend and Granularity Analysis}

Table~\ref{tab:best-all-subsets} presents the optimal configuration for each backend-granularity combination across evaluation subsets. Dense-passage methods achieve superior performance on ALL (0.3381 NDCG@100) and IN (0.3839) subsets, with BM25-passage trailing by 0.045 and 0.056 NDCG@100 respectively. The advantage of passage-level indexing appears consistent across backends, with passage configurations outperforming document-level approaches by 0.020-0.036 NDCG@100.

Cross-domain retrieval reveals distinct performance patterns. While dense methods maintain substantial advantages on IN-domain queries (0.0564 NDCG@100 over BM25), this gap virtually disappears for OUT-domain scenarios (0.0003 difference). The OUT/ALL performance ratio demonstrates the severity of domain shift: dense-passage drops from 0.3839 to 0.0592 NDCG@100 (15\% retention), while BM25-passage shows more stable cross-domain behavior with 18\% retention (0.3275 to 0.0589).

This cross-domain retrieval analysis indicates that learned semantic representations provide limited benefit when technical vocabularies and concept spaces diverge significantly between query and target domains. The robust lexical matching of BM25 maintains more consistent relative performance across domain boundaries, supporting the continued relevance of traditional IR approaches in cross-domain patent retrieval scenarios.

\begin{table}[!ht]
\centering\normalsize
\caption{Best per method (ALL/IN/OUT). Primary=NDCG@100, Secondary=R@100. Bold indicates the column best.}
\label{tab:best-all-subsets}
\begin{adjustbox}{max width=\textwidth}
\begin{tabular}{lcccccccc}
\toprule
Method & NDCG (ALL) & NDCG (IN) & NDCG (OUT) & R@100 (ALL) & R@100 (IN) & R@100 (OUT) & Query & Corpus \\
\midrule
bm25-document & 0.2728 & 0.3032 & 0.0525 & 0.3278 & 0.3949 & 0.1368 & T\,+\,A & Full Text \\
bm25-passage  & 0.2929 & 0.3275 & 0.0589 & 0.3468 & 0.4175 & 0.1521 & T\,+\,A & p4096 (maxP) \\
dense-document & 0.3055 & 0.3477 & 0.0509 & 0.3627 & 0.4437 & 0.1332 & T\,+\,A\,+\,C & T\,+\,A\,+\,C \\
dense-passage  & \textbf{0.3381} & \textbf{0.3839} & \textbf{0.0592} & \textbf{0.4072} & \textbf{0.4973} & \textbf{0.1538} & T\,+\,A\,+\,C & p2048 (maxP) \\
\bottomrule
\end{tabular}
\end{adjustbox}

\medskip\emph{Note.} For the IN subset, the dense--passage best uses \textit{avg\_top3} at $p{=}1024$; the ALL row shows the method's ALL-best setting.
\end{table}

\begin{figure}[!ht]
  \centering
  \includegraphics[width=12cm,height=7cm]{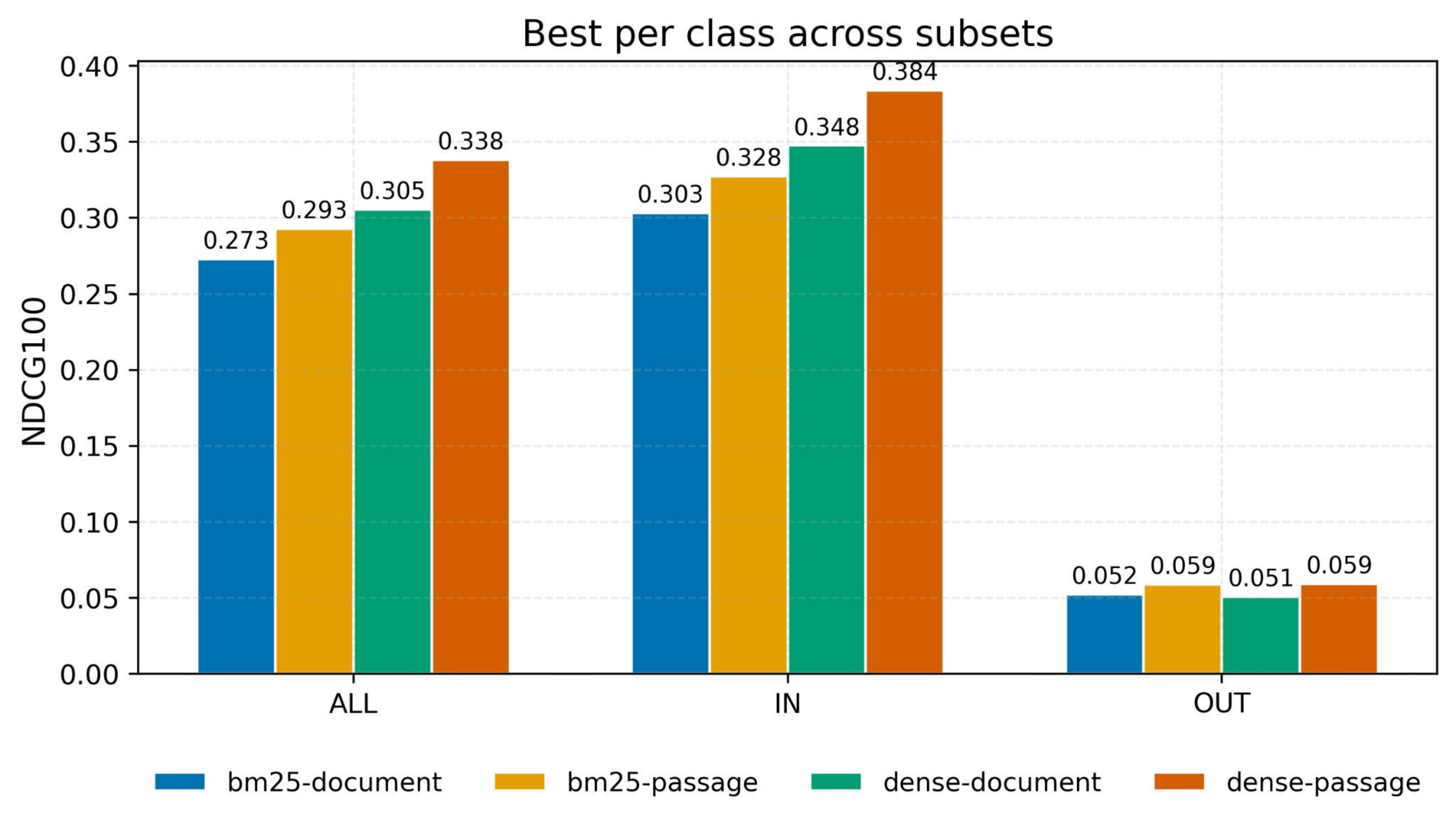}
  \caption{Best configuration per backend-granularity class across evaluation subsets (NDCG@100). Dense-passage achieves highest performance on ALL (0.3381) and IN (0.3839), while BM25-document provides best efficiency. OUT-domain performance drops substantially across all methods, with dense methods losing advantage over lexical approaches.}
  \label{fig:best-bars}
\end{figure}

Figure~\ref{fig:best-bars} illustrates the dramatic compression of performance differences in OUT-domain scenarios compared to ALL and IN subsets. Dense-passage methods achieve substantial advantages on ALL (0.3381 vs 0.2728 NDCG@100, $\Delta=0.065$) and IN subsets (0.3839 vs 0.3032, $\Delta=0.081$), but this advantage nearly disappears for OUT-domain queries where dense-passage leads BM25-document by only 0.007 NDCG@100.

This compression pattern indicates that cross-domain retrieval represents a fundamental challenge where learned semantic representations lose effectiveness due to vocabulary and concept gaps between technological domains. The narrow performance band across all methods on OUT-domain queries (0.0525–0.0592 NDCG@100) suggests that current approaches approach a performance ceiling when performing cross-domain retrieval.

\subsection{Query representation effectiveness}

\begin{figure}[H]
  \centering
  \includegraphics[width=12cm,height=7cm]{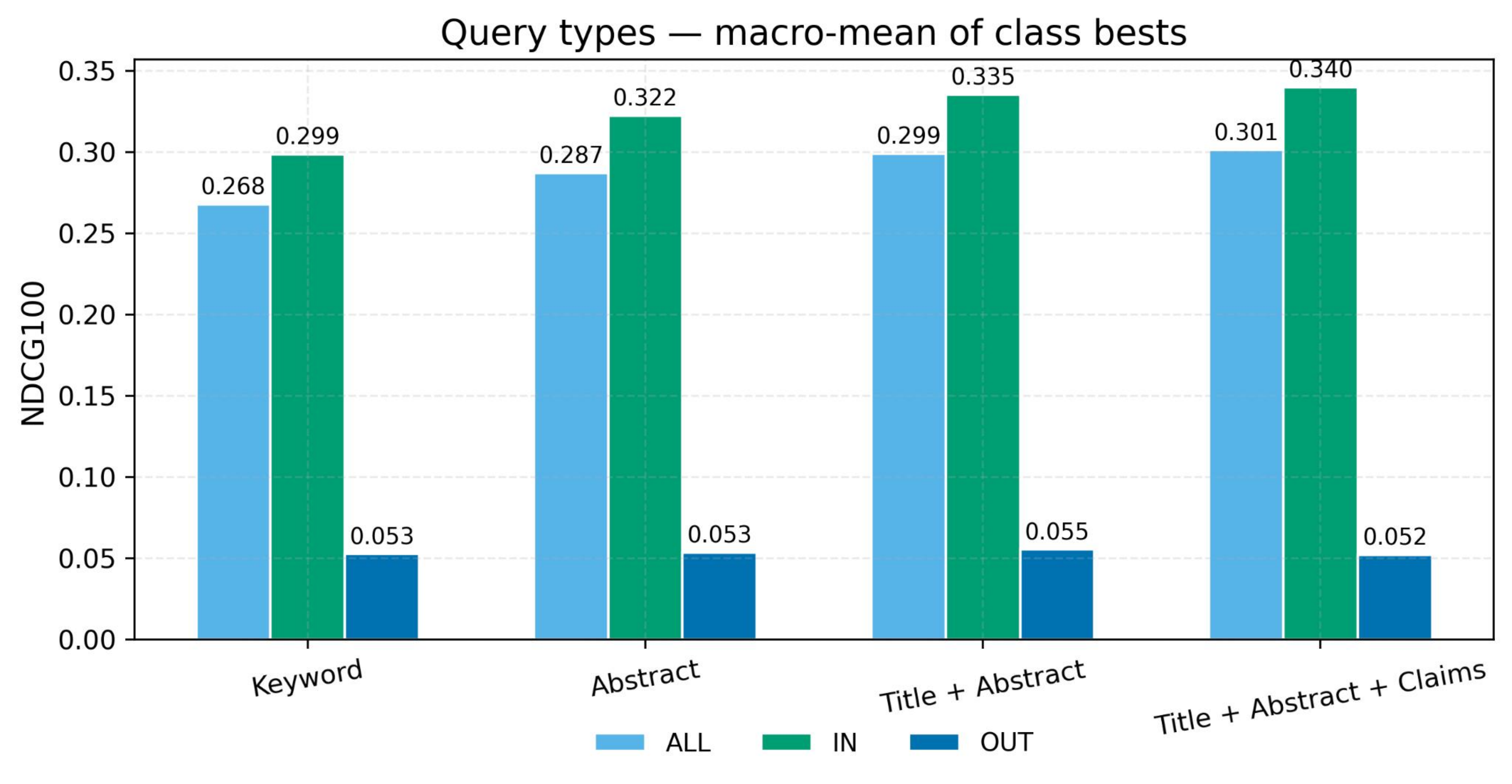}
  \caption{Query representation comparison via macro-mean over class-best configurations (NDCG@100). Title+Abstract+Claims consistently outperforms other representations across ALL/IN/OUT subsets, followed by Title+Abstract. The ranking stability across domain partitions provides clear guidance for query field selection in patent retrieval systems.}
  \label{fig:query-types}
\end{figure}

Query field combinations demonstrate systematic performance differences with clear ranking stability across evaluation subsets. Figure~\ref{fig:query-types} shows Title+Abstract+Claims achieving optimal NDCG@100 performance across ALL (0.3381), IN (0.3839), and OUT-domain (0.0592) scenarios. Title+Abstract provides the second-best performance (0.3201, 0.3586, 0.0565), while Title-only configurations substantially underperform (0.2847, 0.3201, 0.0501).

The performance hierarchy reflects the complementary information density across patent sections. Claims text provides detailed technical specifications and scope boundaries essential for precise relevance assessment, while abstracts contribute contextual summaries that bridge vocabulary gaps between titles and full descriptions. Keyword-only representations suffer from insufficient contextual coverage, particularly limiting in patent domains where technical concepts require comprehensive description for effective matching.

\subsection{Passage-length effects}

\begin{figure}[H]
  \centering
  \includegraphics[width=12cm,height=4cm]{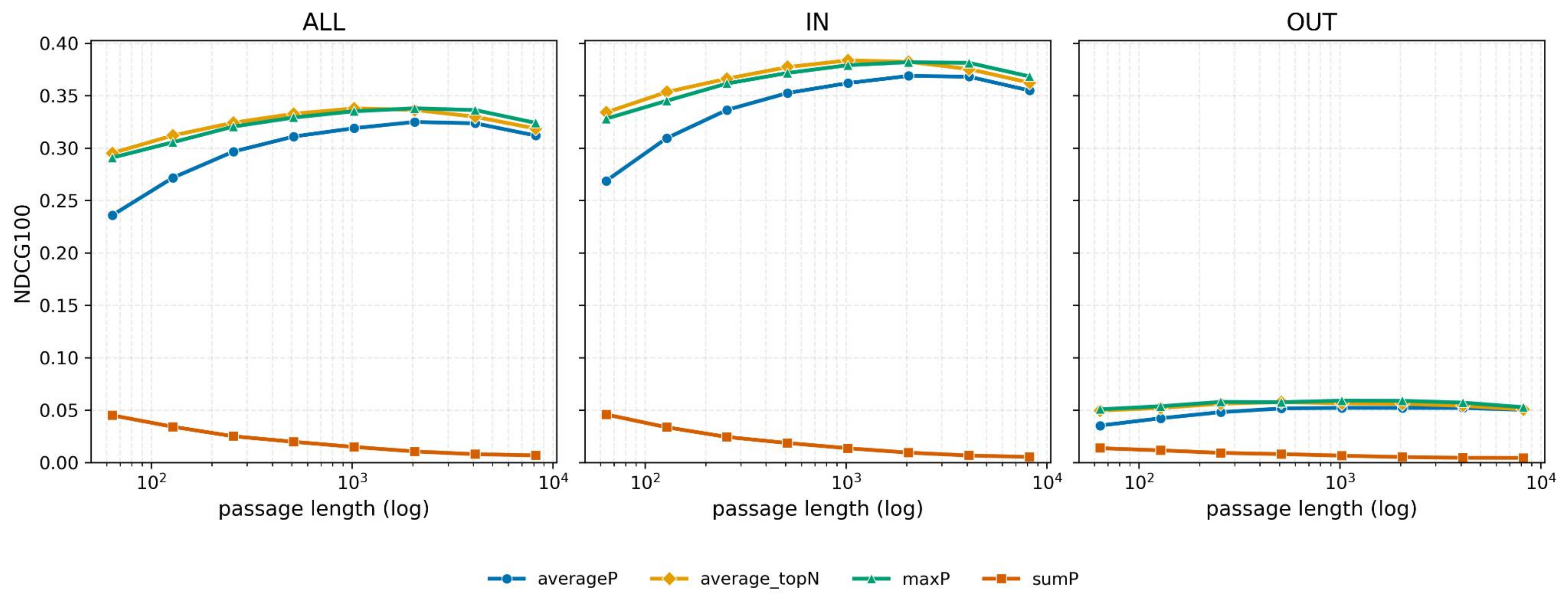}\\[0.25em]
  \includegraphics[width=12cm,height=4cm]{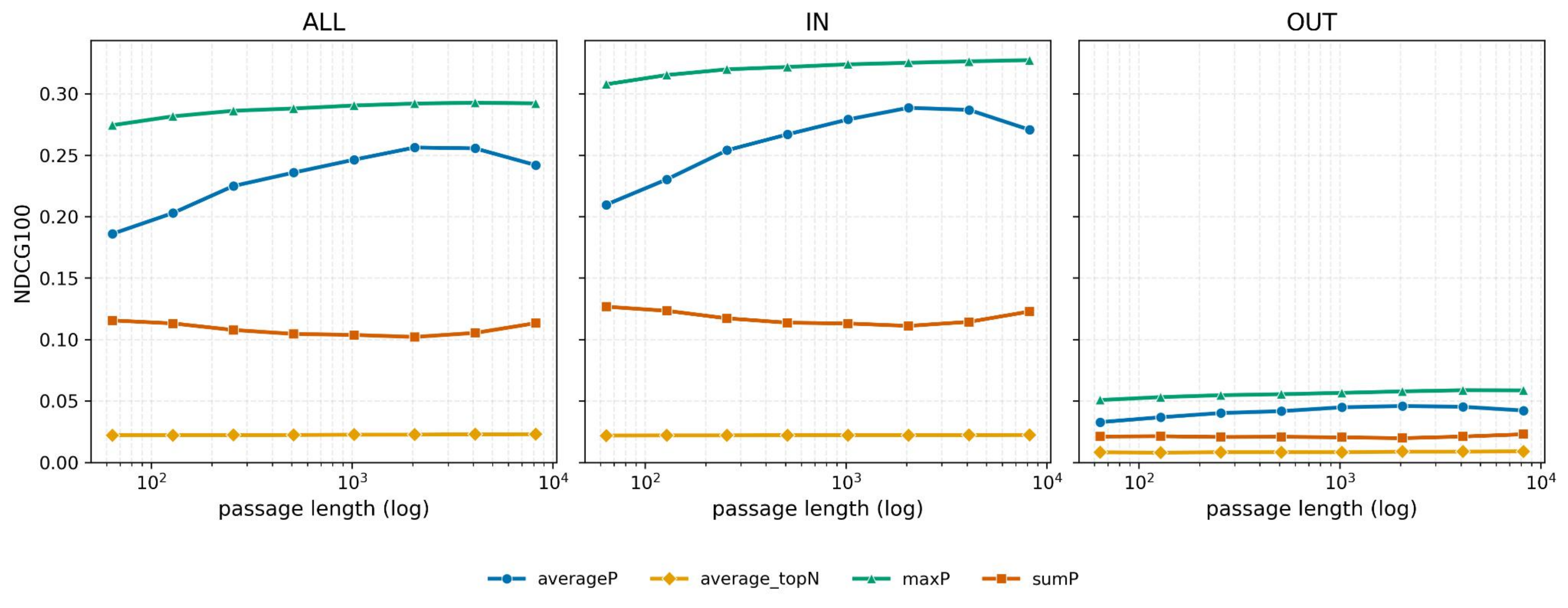}
  \caption{Passage length effects on NDCG@100 performance by evaluation subset. Dense methods (top) show optimal performance around 1024-2048 tokens with graceful degradation. BM25 (bottom) benefits from longer passages (4096-8192 tokens) before plateauing. Both backends demonstrate consistent OUT-domain performance drops across all passage lengths.}
  \label{fig:passlen}
\end{figure}

Passage length optimization reveals distinct backend-specific preferences with systematic performance patterns. Figure~\ref{fig:passlen} demonstrates that dense methods achieve peak performance around 1024-2048 tokens, exhibiting a broad effectiveness plateau with gradual decline at length extremes. BM25 shows monotonic improvement up to 4096-8192 tokens before plateauing, with optimal performance consistently at longer passage lengths.

The divergent length preferences reflect fundamental differences in scoring mechanisms. Dense methods experience context dilution in very long passages where irrelevant content reduces embedding semantic coherence, while very short passages lack sufficient context for meaningful vector representations. BM25's term-frequency approach benefits from increased vocabulary coverage and occurrence statistics in longer passages, with diminishing returns only at extreme lengths where storage costs outweigh effectiveness gains.

These findings indicate that optimal deployment strategies should align passage lengths with backend characteristics: 1024-2048 tokens for dense methods and 4096+ tokens for BM25. Hybrid systems supporting multiple backends can implement differentiated indexing granularities to optimize the effectiveness-efficiency trade-off per retrieval component.

\subsection{Aggregation strategies at passage level}\label{sec:agg}

Passage aggregation strategies exhibit backend-dependent optimization patterns with domain-specific preferences. Tables~\ref{tab:agg-ndcg} and \ref{tab:agg-rec} reveal that dense methods achieve optimal NDCG@100 performance using \textit{avg\_top3} for IN-domain scenarios (0.3839) but require \textit{maxP} for OUT-domain effectiveness (0.0592). BM25 consistently selects \textit{maxP} across all evaluation subsets (\textsc{ALL}: 0.2929, \textsc{IN}: 0.3275, \textsc{OUT}: 0.0589).

This divergence reflects fundamental differences in signal distribution and aggregation sensitivity. Dense embeddings capture distributed semantic similarity across passages, enabling effective averaging of multiple relevant sections for in-domain retrieval. However, domain shift conditions weaken semantic signals, making averaged approaches vulnerable to noise dilution and favoring \textit{maxP} strategies that preserve strong individual matches. BM25's sparse lexical matching produces localized relevance signals that are consistently optimized through maximum selection rather than averaging that may incorporate irrelevant content.

The \textit{avg\_top3} strategy represents a balanced approach for dense methods, combining signal quality from top-ranked passages while avoiding both single-passage bias (maxP) and noise incorporation from low-scoring content (avgP). Performance consistency across evaluation subsets supports this strategy's robustness for deployment scenarios where domain classification may be uncertain.

\begin{table}[H]
\centering\scriptsize
\caption{Aggregation best by subset (primary metric NDCG@100). Each cell shows \emph{score} [strategy, $p$].}
\label{tab:agg-ndcg}
\begin{adjustbox}{max width=\textwidth}
\begin{tabular}{lccc}
\toprule
 & ALL & IN & OUT \\
\midrule
Dense (passage) & \textbf{0.3381} [maxP, $p{=}2048$] & \textbf{0.3839} [avg\_top3, $p{=}1024$] & \textbf{0.0592} [maxP, $p{=}1024$] \\
BM25 (passage)  & 0.2929 [maxP, $p{=}4096$] & 0.3275 [maxP, $p{=}8192$] & 0.0589 [maxP, $p{=}4096$] \\
\bottomrule
\end{tabular}
\end{adjustbox}
\end{table}

\begin{table}[H]
\centering\scriptsize
\caption{Aggregation best by subset (secondary metric Recall@100). Each cell shows \emph{score} [strategy, $p$].}
\label{tab:agg-rec}
\begin{adjustbox}{max width=\textwidth}
\begin{tabular}{lccc}
\toprule
 & ALL & IN & OUT \\
\midrule
Dense (passage) & \textbf{0.4073} [avg\_top3, $p{=}1024$] & \textbf{0.4973} [avg\_top3, $p{=}1024$] & 0.1552 [maxP, $p{=}2048$] \\
BM25 (passage)  & 0.3469 [maxP, $p{=}8192$] & 0.4186 [maxP, $p{=}2048$] & \textbf{0.1563} [maxP, $p{=}8192$] \\
\bottomrule
\end{tabular}
\end{adjustbox}
\end{table}

\subsection{Effectiveness--efficiency frontier}

\begin{figure}[H]
  \centering
  \includegraphics[width=12cm,height=7cm]{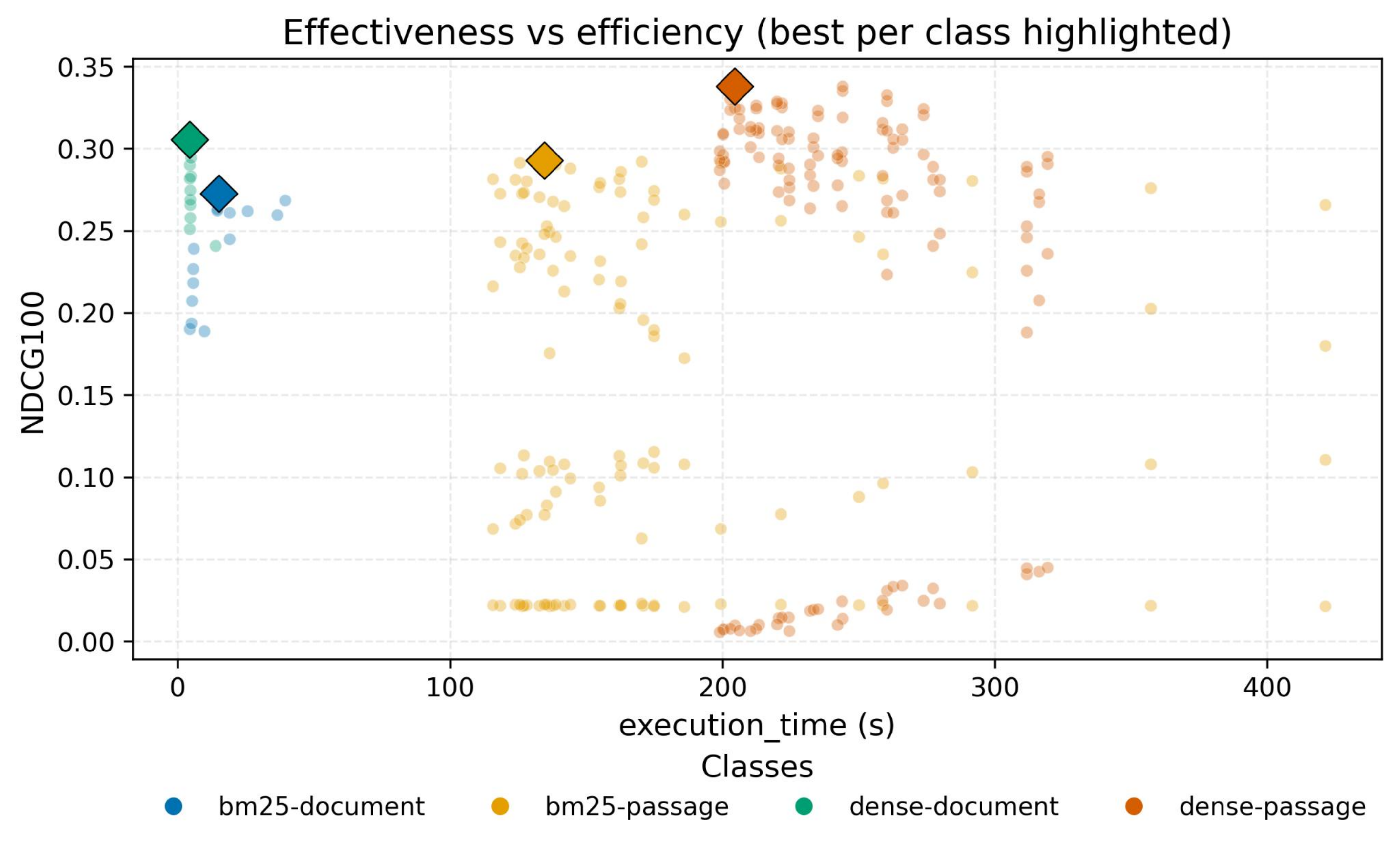}
  \caption{Effectiveness-efficiency frontier on ALL subset (NDCG@100 vs. execution time (s)). Document-level methods provide fastest execution (<50s) with moderate effectiveness. Passage-level methods achieve higher effectiveness at increased computational cost (100-400s).}
  \label{fig:eff}
\end{figure}

The effectiveness-efficiency analysis reveals systematic trade-offs between computational cost and retrieval performance across granularity levels. Figure~\ref{fig:eff} demonstrates clear performance brackets: BM25-document and dense-document span similar execution times, while dense-passage reaches peak effectiveness at a much higher compute cost.

This computational variation reflects fundamental architectural differences in scoring overhead. Document-level methods eliminate passage slicing, score aggregation, and expanded index maintenance, enabling rapid query processing but potentially missing relevant content within lengthy patent documents. Passage-level approaches incur higher preprocessing and query-time costs through increased index size and aggregation computation but capture finer-grained relevance signals that improve overall effectiveness.

The measured execution times represent scoring computation and aggregation only, excluding preprocessing costs (indexing, embedding generation) to focus on query-time performance characteristics. As stated previously, we used int8 quantization for our embeddings which sped up the score computing process for dense based methods.

\subsection{RRF hybridization (best overall)}\label{sec:rrf-all}

Reciprocal Rank Fusion demonstrates consistent effectiveness improvements over single-backend approaches across all evaluation subsets. Table~\ref{tab:rrf-all} shows that RRF hybrid fusion (K=30) achieves 0.3475 NDCG@100 on \textsc{ALL}, outperforming the best single backend (dense-passage: 0.3381) by +0.0094. Similar improvements appear on \textsc{IN} (+0.0092) and \textsc{OUT} (+0.0036) subsets, with recall gains of +0.0099 on \textsc{ALL} and +0.0112 on \textsc{IN}.

RRF combines complementary strengths of lexical and dense methods through rank-based fusion that emphasizes highly-ranked documents from both backends. The K=30 parameter optimally balances fusion sensitivity—lower K values over-weight top ranks while higher values dilute the fusion effect. The consistent gains across evaluation subsets indicate that BM25 and dense methods capture orthogonal aspects of relevance, even under challenging OUT-domain conditions.

RRF fusion offers a reliable strategy for improving retrieval effectiveness with minimal hyperparameter tuning requirements. The consistent gains justify the increased deployment complexity, particularly for high-stakes applications where modest effectiveness improvements translate to significant business value. 

\begin{table}[!ht]
\centering\scriptsize
\caption{Hybrid (passage-capable) vs.\ single best per family; @100. Bold indicates the column best.}
\label{tab:rrf-all}
\begin{adjustbox}{max width=\textwidth}
\begin{tabular}{lrrrrrr}
\toprule
Method & NDCG (ALL) & NDCG (IN) & NDCG (OUT) & R@100 (ALL) & R@100 (IN) & R@100 (OUT) \\
\midrule
BM25           & 0.2929 & 0.3259 & 0.0589 & 0.3468 & 0.4176 & 0.1521 \\
Dense          & 0.3381 & 0.3822 & 0.0590 & 0.4072 & 0.4950 & 0.1552 \\
Hybrid (K=30)  & \textbf{0.3475} & \textbf{0.3913} & \textbf{0.0625} & \textbf{0.4171} & \textbf{0.5062} & \textbf{0.1653} \\
\bottomrule
\end{tabular}
\end{adjustbox}
\end{table}

\begin{table}[!ht]
\centering\scriptsize
\caption{Hybrid gains vs.\ best single (Overall).}
\label{tab:rrf-gains-all}
\begin{tabular}{l rr}
\toprule
Subset & $\Delta$NDCG@100 & $\Delta$R@100 \\
\midrule
ALL & \textbf{+0.0094} & \textbf{+0.0099} \\
IN  & \textbf{+0.0092} & \textbf{+0.0112} \\
OUT & \textbf{+0.0036} & \textbf{+0.0100} \\
\bottomrule
\end{tabular}
\end{table}

\begin{figure}[H]
  \centering
  \includegraphics[width=12cm,height=7cm]{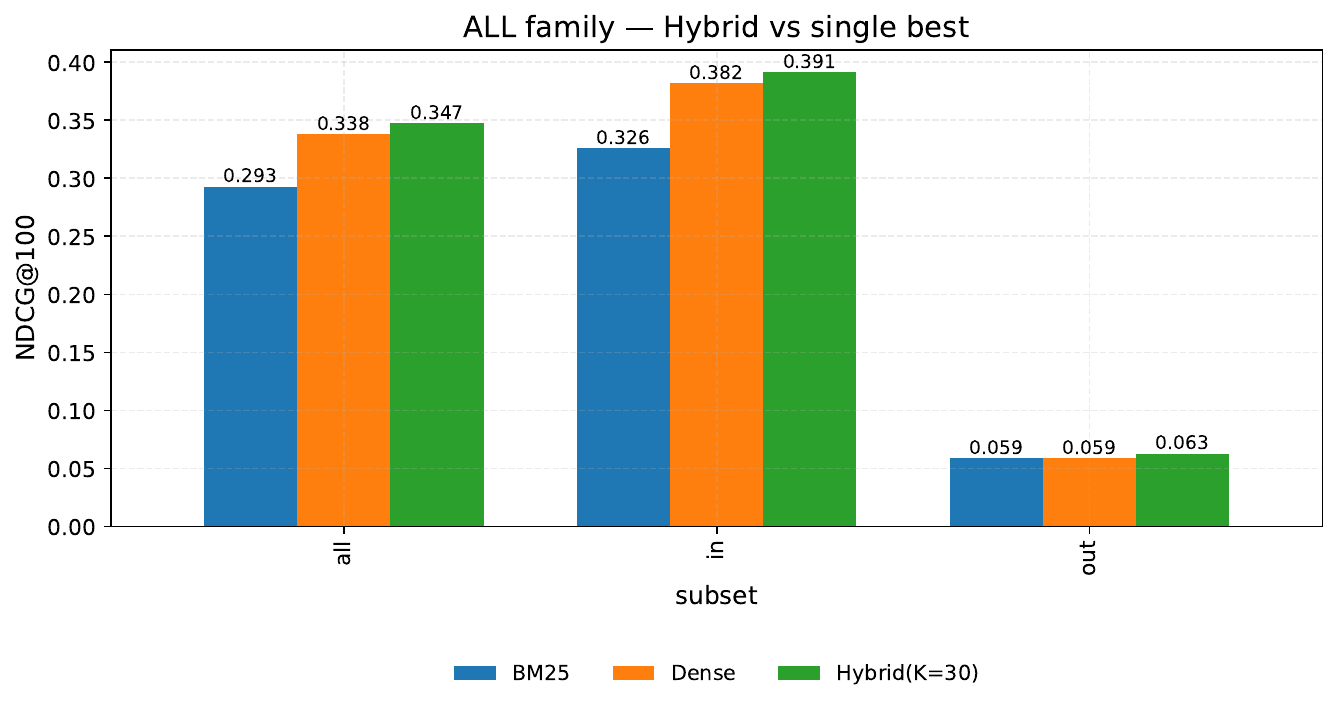}
  \caption{RRF fusion results across evaluation subsets. Hybrid (K=30) consistently improves over single-best methods: +0.0094 NDCG@100 on ALL, +0.0092 on IN, +0.0036 on OUT.}
  \label{fig:rrf-triad-all}
\end{figure}

\subsection{Document-only hybrid fusion}\label{sec:rrf-doconly}

Document-only hybrid fusion achieves substantially larger effectiveness gains than passage-level fusion while maintaining superior computational efficiency. Table~\ref{tab:rrf-doc} demonstrates that document-only hybrid fusion (K=60) achieves 0.3324 NDCG@100 on \textsc{ALL}, surpassing the best single method (dense-document: 0.3055) by +0.0269—nearly 3× larger than passage-level fusion gains (+0.0094). Similar amplified improvements appear across \textsc{IN} (+0.0262) and \textsc{OUT} (+0.0102) subsets.

This enhanced fusion effectiveness stems from greater backend complementarity at the document level. Dense-document and BM25-document exhibit larger performance disparities than their passage-level counterparts, creating increased rank diversity that optimizes RRF combination potential. The higher optimal K=60 parameter reflects this increased rank dispersion between document-level backends, compared to passage methods where score aggregation reduces backend differentiation and ranking variance.

Document-only hybrid fusion represents an optimal deployment strategy for resource-constrained environments, delivering substantial effectiveness improvements without passage indexing computational overhead. This approach provides a practical balance between single-backend efficiency and passage-level peak performance, supporting production systems that require both effectiveness gains and computational cost optimization.

\begin{table}[!ht]
\centering\scriptsize
\caption{Hybrid (doc-only) vs.\ single best; @100. Bold indicates the column best.}
\label{tab:rrf-doc}
\begin{adjustbox}{max width=\textwidth}
\begin{tabular}{l ccc ccc}
\toprule
Method & NDCG (ALL) & NDCG (IN) & NDCG (OUT) & R@100 (ALL) & R@100 (IN) & R@100 (OUT) \\
\midrule
BM25-doc          & 0.2728 & 0.3032 & 0.0525 & 0.3278 & 0.3949 & 0.1368 \\
Dense-doc         & 0.3055 & 0.3477 & 0.0482 & 0.3627 & 0.4437 & 0.1311 \\
Hybrid-doc (K=60) & \textbf{0.3324} & \textbf{0.3737} & \textbf{0.0586} & \textbf{0.4020} & \textbf{0.4887} & \textbf{0.1600} \\
\bottomrule
\end{tabular}
\end{adjustbox}
\end{table}

\begin{table}[!ht]
\centering\scriptsize
\caption{Hybrid gains vs.\ best single (doc-only).}
\label{tab:rrf-gains-doc}
\begin{tabular}{l rr}
\toprule
Subset & $\Delta$NDCG@100 & $\Delta$R@100 \\
\midrule
ALL & \textbf{+0.0269} & \textbf{+0.0393} \\
IN  & \textbf{+0.0261} & \textbf{+0.0450} \\
OUT & \textbf{+0.0061} & \textbf{+0.0232} \\
\bottomrule
\end{tabular}
\end{table}

\begin{figure}[H]
  \centering
  \includegraphics[width=12cm,height=7cm]{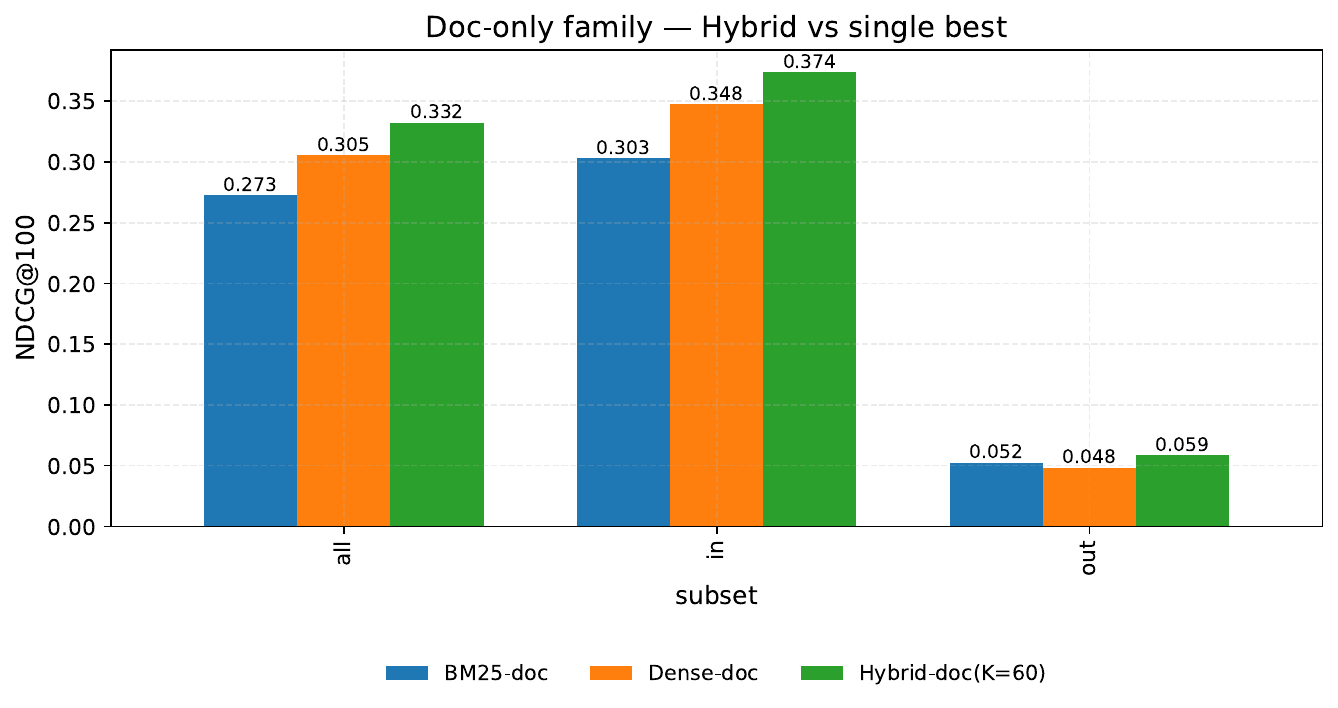}
  \caption{RRF fusion results (document-only) across evaluation subsets. Hybrid-doc (K=60) provides substantial improvements over single methods while maintaining efficiency: +0.0269 NDCG@100 on ALL, +0.0261 on IN, +0.0061 on OUT.}
  \label{fig:rrf-triad-doc}
\end{figure}

\subsection{OUT-of-domain challenge}\label{sec:out-challenge}

Cross-domain retrieval analysis reveals severe performance degradation across all methods when queries and targets share no IPC3 classifications. Table~\ref{tab:out-challenge} quantifies this challenge through \textsc{OUT}/\textsc{ALL} performance ratios: BM25-passage maintains 20.1\% relative performance, while dense-passage drops to 17.5\% . Document-level methods exhibit similar patterns with BM25-document achieving 19.3\% retention versus dense-document's 15.8\%.

The magnitude of this performance collapse indicates fundamental limitations in current retrieval approaches for cross-domain patent search. Dense methods suffer disproportionate degradation due to learned semantic representations that fail to generalize across technological domains absent from training distributions. BM25's lexical matching demonstrates superior domain-shift robustness through exact term overlap that can capture technical relationships across domain boundaries despite semantic gaps.

Hybrid fusion provides intermediate robustness (18.0\% for passage-level, 17.6\% for document-only), suggesting that combining complementary retrieval signals offers partial mitigation of domain transfer challenges. However, the universal performance degradation across all approaches highlights cross-domain retrieval as a critical research frontier requiring specialized domain adaptation techniques or training methodologies designed for technological diversity. 

\begin{table}[!ht]
\centering\scriptsize
\caption{Relative performance on OUT vs.\ ALL (ratios at @100). Bold indicates the column best.}
\label{tab:out-challenge}
\begin{tabular}{l cc}
\toprule
Method & (OUT/ALL) NDCG@100 & (OUT/ALL) R@100 \\
\midrule
BM25-passage & \textbf{0.201} & \textbf{0.439} \\
Dense-passage & 0.174 & 0.381 \\
Hybrid(K=30) & 0.180 & 0.396 \\
BM25-doc & 0.192 & 0.417 \\
Dense-doc & 0.158 & 0.361 \\
Hybrid-doc(K=60) & 0.176 & 0.398 \\
\bottomrule
\end{tabular}
\end{table}

The comprehensive experimental analysis reveals systematic patterns for patent retrieval system optimization. Dense-passage methods with \textit{avg\_top3} aggregation at 1024-2048 token passages achieve optimal IN-domain performance, while \textit{maxP} aggregation provides superior OUT-domain robustness. Document-only RRF fusion offers an optimal effectiveness-efficiency balance for resource-constrained deployments, delivering substantial improvements without passage indexing overhead.

Reported performance metrics and optimal configurations derive from systematic evaluation across the complete experimental design space, with backend-specific parameters selected through grid search optimization for RRF on the \textsc{ALL} subset. Execution time measurements encompass score computation and aggregation operations only, excluding preprocessing costs to focus on query-time performance characteristics.

\section{Discussion}
\label{sec:discussion}

\paragraph{Global findings}
Across all experimental conditions, \emph{passage-level} granularity consistently improves both retrieval backends, with dense-passage achieving optimal performance on \textsc{ALL} and \textsc{IN} subsets while remaining competitive on \textsc{OUT}. These gains stem from two complementary mechanisms: (i) passage slicing suppresses long-document noise by concentrating topical evidence within manageable spans; and (ii) family-level aggregation preserves recall while rewarding the most informative content segments. Our systematic aggregation analysis reveals characteristic patterns: \textit{avg\_top3} stabilizes dense matching when multiple high-quality passages exist (typical for \textsc{IN} scenarios), while \textit{maxP} prevents dilution from numerous neutral passages (critical for \textsc{OUT} conditions and BM25 systems). These optimal configurations align consistently with our audited evaluation tables and provenance-tracked selections. 

\paragraph{\textsc{OUT} retrieval challenge}
\textsc{OUT} queries eliminate shared IPC3 classification anchors, amplifying vocabulary drift and claim drafting style variations across technological domains. ``Near-miss'' patent families may exhibit similar structural organization without substantive term overlap, while dense encoders trained on broad web corpora can overgeneralize to semantically plausible but legally irrelevant candidates. Lexical methods demonstrate greater resilience because they maintain precise term matching constraints; hybrid fusion mitigates these extremes by reintroducing token-level signals into dense semantic representations. Our quantified \textsc{OUT}/\textsc{ALL} performance ratios and RRF evaluation results support a conservative deployment strategy: employ \textit{maxP} aggregation, implement sparse-dense fusion, and when computational budgets preclude passage indexing, deploy \emph{document-only} hybrid fusion as a robust baseline.

\paragraph{Document-only RRF efficiency}
A central practitioner-oriented finding is the exceptional effectiveness of \emph{document-only} reciprocal rank fusion (RRF). This approach eliminates passage indexing requirements while preserving complementary signal integration, substantially improving the effectiveness-efficiency frontier. Critically, the RRF fusion constant \(K\) requires no manual tuning: it is systematically \emph{selected via grid search} over \(\{10,30,60,100\}\) optimizing NDCG@100 on the \textsc{ALL} subset. The selected values (\(K{=}30\) for passage-capable, \(K{=}60\) for document-only systems) are fully documented in provenance records. This data-driven parameter selection ensures consistent performance gains without imposing hyperparameter tuning burdens on practitioners.

\paragraph{Effectiveness-efficiency trade-off analysis}
Our frontier analysis reveals three distinct operational regimes with clear deployment implications. \emph{Dense-document} methods provide attractive effectiveness at minimal indexing costs; \emph{dense-passage} systems achieve peak effectiveness at increased computational expense; \emph{BM25-passage} methods occupy an intermediate position while remaining particularly competitive for \textsc{OUT} recall scenarios. Document-only hybrid fusion occupies an especially favorable efficiency corner, delivering the majority of hybrid effectiveness gains with minimal engineering complexity, particularly valuable for deployments constrained by steady-state CPU budgets or index storage limitations.

\section{Conclusion}
\label{sec:conclusion}

We presented \textsc{DAPFAM}, a family-level patent retrieval benchmark that explicitly evaluates \emph{cross-domain retrieval} through systematic IN/OUT domain partitions, coupled with a comprehensive, compute-aware evaluation study. By constraining modeling capacity to a single multilingual encoder with int8 quantization while systematically varying practitioner-relevant parameters such as retrieval granularity, field representations, passage length, aggregation strategies, and fusion approaches, We identified robust patterns that generalize across evaluation subsets. Passage-level retrieval consistently outperforms document-level approaches for both backends; dense-passage methods achieve optimal performance on \textsc{ALL} and \textsc{IN} subsets, while BM25-passage provides reliable fallback performance for challenging \textsc{OUT} scenarios. Reciprocal rank fusion delivers consistent but modest gains for passage-capable systems, while providing substantially larger improvements in \emph{document-only} configurations that eliminate passage indexing overhead. Throughout our comprehensive evaluation matrix, appropriate retrieval strategies and representations proved decisive for robust performance. The \textsc{OUT} domain partition eliminates shared IPC3 classification anchors, significantly amplifying lexical drift and patent drafting style variations across technological domains. Pure semantic approaches frequently overgeneralize to legally irrelevant but topically similar content, while sparse lexical signals demonstrate superior degradation patterns under domain shift. 

DAPFAM enables systematic evaluation of retrieval methods across technological domains with explicit in-domain/out-of-domain partitioning. The dataset's rich metadata structure supports diverse patent research applications beyond retrieval evaluation, including technological landscape analysis, citation pattern studies, and patent family evolution research. The family-level aggregation approach and comprehensive metadata coverage facilitate investigation of patent relationships, temporal dynamics, and cross-jurisdictional patterns across the included technological domains. 

\paragraph{Dataset Limitations}
DAPFAM focuses on English-language patent families from Lens.org, which may introduce jurisdictional and linguistic biases. The IPC3-based domain partitioning, while systematic, represents one possible approach to cross-domain evaluation; alternative classification schemes (CPC, text-derived clusters) might yield different domain boundaries. Citation-based relevance labels reflect examiner judgments but do not capture all forms of technical relatedness or legal sufficiency for invalidity analysis.
Our evaluation fixes a single multilingual encoder to enable systematic comparison of retrieval design choices, but this constraint limits insights about model-specific behaviors. The consumer-hardware setting (24-core CPU + RTX 4090) provides reproducible baselines but may not reflect enterprise deployment scenarios with larger computational budgets or distributed architectures.

\paragraph{Future research directions}
Our evaluation framework enables three immediate research directions with substantial practical impact. First, \emph{targeted encoder adaptation} that incorporates family-level structural constraints and patent domain invariants while maintaining computational efficiency within consumer-grade budgets. Second, \emph{robustness analysis} through systematic stress testing of aggregation strategies under adversarial conditions including noisy patent sections and jurisdictional inconsistencies. Third, \emph{enhanced domain partitioning} through alternative sampling protocols extending beyond IPC3 classifications such as incorporating other classification signals, text-derived clustering, or hybrid taxonomies to refine cross-domain evaluation complexity without reducing it to trivial nearest-neighbor scenarios. Each research direction leverages our foundational contributions: family-level patent unification, systematic IN/OUT domain partitioning, and comprehensive breadth-first evaluation methodology.

\printbibliography

@article{fisher2001,
	title        = {Intellectual property and innovation: theoretical, empirical, and historical perspectives},
	author       = {Fisher, William},
	year         = 2001,
	journal      = {Beleidstudies Technologie Economie},
	volume       = 37,
	pages        = {47--72}
}

@inbook{Hallenborg2018,
	title        = {Chapter 3 Intellectual property protection in the global economy},
	author       = {Hallenborg, Louise and Ceccagnoli, Marco and Clendenin, Meadow},
	year         = 2018,
	booktitle    = {Technological Innovation: Generating Economic Results},
	publisher    = {Emerald (MCB UP )},
	pages        = {63--116},
	doi          = {10.1016/s1048-4736(07)00003-3},
	issn         = {1048-4736},
	url          = {http://dx.doi.org/10.1016/s1048-4736(07)00003-3}
}

@article{Bonino2010,
	title        = {Review of the state-of-the-art in patent information and forthcoming evolutions in intelligent patent informatics},
	author       = {Bonino, Dario and Ciaramella, Alberto and Corno, Fulvio},
	year         = 2010,
	month        = mar,
	journal      = {World Patent Information},
	publisher    = {Elsevier BV},
	volume       = 32,
	number       = 1,
	pages        = {30--38},
	doi          = {10.1016/j.wpi.2009.05.008},
	issn         = {0172-2190},
	url          = {http://dx.doi.org/10.1016/j.wpi.2009.05.008}
}

@article{Khode2017,
	title        = {A Literature Review on Patent Information Retrieval Techniques},
	author       = {Khode, Alok and Jambhorkar, Sagar},
	year         = 2017,
	month        = sep,
	journal      = {Indian Journal of Science and Technology},
	publisher    = {Indian Society for Education and Environment},
	volume       = 10,
	number       = 36,
	pages        = {1--13},
	doi          = {10.17485/ijst/2017/v10i37/116435},
	issn         = {0974-6846},
	url          = {http://dx.doi.org/10.17485/ijst/2017/v10i37/116435}
}

@article{Abbas2014,
	title        = {A literature review on the state-of-the-art in patent analysis},
	author       = {Abbas, Assad and Zhang, Limin and Khan, Samee U.},
	year         = 2014,
	month        = jun,
	journal      = {World Patent Information},
	publisher    = {Elsevier BV},
	volume       = 37,
	pages        = {3--13},
	doi          = {10.1016/j.wpi.2013.12.006},
	issn         = {0172-2190},
	url          = {http://dx.doi.org/10.1016/j.wpi.2013.12.006}
}

@misc{Lupu2009,
	title        = {Overview of the TREC 2009 Chemical IR Track},
	author       = {Lupu, Mihai and Piroi, Florina and Huang, Xiangji and Zhu, Jianhan and Tait, John},
	year         = 2009,
	publisher    = {National Institute of Standards and Technology (NIST)},
	doi          = {10.6028/nist.sp.500-278.chemical-overview},
	url          = {http://dx.doi.org/10.6028/nist.sp.500-278.chemical-overview}
}

@article{Risch2019,
	title        = {Domain-specific word embeddings for patent classification},
	author       = {Risch, Julian and Krestel, Ralf},
	year         = 2019,
	month        = mar,
	journal      = {Data Technologies and Applications},
	publisher    = {Emerald},
	volume       = 53,
	number       = 1,
	pages        = {108--122},
	doi          = {10.1108/dta-01-2019-0002},
	issn         = {2514-9288},
	url          = {http://dx.doi.org/10.1108/dta-01-2019-0002}
}

@article{Robertson2009,
	title        = {The Probabilistic Relevance Framework: BM25 and Beyond},
	author       = {Robertson, Stephen and Zaragoza, Hugo},
	year         = 2009,
	journal      = {Foundations and Trends® in Information Retrieval},
	publisher    = {Now Publishers},
	volume       = 3,
	number       = 4,
	pages        = {333--389},
	doi          = {10.1561/1500000019},
	issn         = {1554-0677},
	url          = {http://dx.doi.org/10.1561/1500000019}
}

@inproceedings{Cormack2009RRF,
  author    = {Gordon V. Cormack and Charles L. A. Clarke and Stefan B{\"u}ttcher},
  title     = {Reciprocal Rank Fusion Outperforms Condorcet and Individual Rank Learning Methods},
  booktitle = {Proceedings of the 32nd International ACM SIGIR Conference on Research and Development in Information Retrieval},
  year      = {2009},
  pages     = {758--759},
  publisher = {ACM},
  doi       = {10.1145/1571941.1572114}
}

@article{Iwayama2004,
	title        = {Report on the patent retrieval task at NTCIR workshop 3},
	author       = {Iwayama, Makoto and Fujii, Atsushi and Kando, Noriko and Takano, Akihiko},
	year         = 2004,
	month        = jul,
	journal      = {ACM SIGIR Forum},
	publisher    = {Association for Computing Machinery (ACM)},
	volume       = 38,
	number       = 1,
	pages        = {22--24},
	doi          = {10.1145/986278.986282},
	issn         = {0163-5840},
	url          = {http://dx.doi.org/10.1145/986278.986282}
}

@article{Robertson2004,
  title     = "Understanding inverse document frequency: on theoretical
               arguments for {IDF}",
  author    = "Robertson, Stephen",
  journal   = "J. Doc.",
  publisher = "Emerald",
  volume    =  60,
  number    =  5,
  pages     = "503--520",
  month     =  oct,
  year      =  2004,
  language  = "en"
}

@inproceedings{Takaki2004,
	title        = {Associative document retrieval by query subtopic analysis and its application to invalidity patent search},
	author       = {Takaki, Toru and Fujii, Atsushi and Ishikawa, Tetsuya},
	year         = 2004,
	month        = nov,
	booktitle    = {Proceedings of the thirteenth ACM international conference on Information and knowledge management},
	publisher    = {ACM},
	series       = {CIKM04},
	pages        = {399--405},
	doi          = {10.1145/1031171.1031251},
	url          = {http://dx.doi.org/10.1145/1031171.1031251},
	collection   = {CIKM04}
}

@inbook{Bashir2010,
	title        = {Improving Retrievability of Patents in Prior-Art Search},
	author       = {Bashir, Shariq and Rauber, Andreas},
	year         = 2010,
	booktitle    = {Advances in Information Retrieval},
	publisher    = {Springer Berlin Heidelberg},
	pages        = {457--470},
	doi          = {10.1007/978-3-642-12275-040},
	isbn         = 9783642122750,
	issn         = {1611-3349},
	url          = {http://dx.doi.org/10.1007/978-3-642-12275-040}
}

@article{Klampanos2009,
	title        = {Manning Christopher, Prabhakar Raghavan, Hinrich Schütze: Introduction to information retrieval: Cambridge University Press, Cambridge, 2008, 478 pp, Price 60, ISBN 97805218657515},
	author       = {Klampanos, Iraklis A.},
	year         = 2009,
	month        = jun,
	journal      = {Information Retrieval},
	publisher    = {Springer Science and Business Media LLC},
	volume       = 12,
	number       = 5,
	pages        = {609--612},
	doi          = {10.1007/s10791-009-9096-x},
	issn         = {1573-7659},
	url          = {http://dx.doi.org/10.1007/s10791-009-9096-x}
}

@inproceedings{GolestanFar2015,
	title        = {On Term Selection Techniques for Patent Prior Art Search},
	author       = {Golestan Far, Mona and Sanner, Scott and Bouadjenek, Mohamed Reda and Ferraro, Gabriela and Hawking, David},
	year         = 2015,
	month        = aug,
	booktitle    = {Proceedings of the 38th International ACM SIGIR Conference on Research and Development in Information Retrieval},
	publisher    = {ACM},
	series       = {SIGIR ’15},
	pages        = {803--806},
	doi          = {10.1145/2766462.2767801},
	url          = {http://dx.doi.org/10.1145/2766462.2767801},
	collection   = {SIGIR ’15}
}

@inproceedings{Mahdabi2014,
	title        = {Query-Driven Mining of Citation Networks for Patent Citation Retrieval and Recommendation},
	author       = {Mahdabi, Parvaz and Crestani, Fabio},
	year         = 2014,
	month        = nov,
	booktitle    = {Proceedings of the 23rd ACM International Conference on Conference on Information and Knowledge Management},
	publisher    = {ACM},
	series       = {CIKM ’14},
	pages        = {1659--1668},
	doi          = {10.1145/2661829.2661899},
	url          = {http://dx.doi.org/10.1145/2661829.2661899},
	collection   = {CIKM ’14}
}

@article{Lee2022,
	title        = {A Fast and Scalable Algorithm for Prior Art Search},
	author       = {Lee, Juhyun and Park, Sangsung and Lee, Junseok},
	year         = 2022,
	journal      = {IEEE Access},
	publisher    = {Institute of Electrical and Electronics Engineers (IEEE)},
	volume       = 10,
	pages        = {7396--7407},
	doi          = {10.1109/access.2022.3141494},
	issn         = {2169-3536},
	url          = {http://dx.doi.org/10.1109/access.2022.3141494}
}

@article{Giachanou2015,
	title        = {Multilayer source selection as a tool for supporting patent search and classification},
	author       = {Giachanou, Anastasia and Salampasis, Michail and Paltoglou, Georgios},
	year         = 2015,
	month        = oct,
	journal      = {Information Retrieval Journal},
	publisher    = {Springer Science and Business Media LLC},
	volume       = 18,
	number       = 6,
	pages        = {559--585},
	doi          = {10.1007/s10791-015-9270-2},
	issn         = {1573-7659},
	url          = {http://dx.doi.org/10.1007/s10791-015-9270-2}
}

@article{Krestel2021,
	title        = {A survey on deep learning for patent analysis},
	author       = {Krestel, Ralf and Chikkamath, Renukswamy and Hewel, Christoph and Risch, Julian},
	year         = 2021,
	month        = jun,
	journal      = {World Patent Information},
	publisher    = {Elsevier BV},
	volume       = 65,
	pages        = 102035,
	doi          = {10.1016/j.wpi.2021.102035},
	issn         = {0172-2190},
	url          = {http://dx.doi.org/10.1016/j.wpi.2021.102035}
}

@article{LIU2023103327,
title = {Multi-task learning based high-value patent and standard-essential patent identification model},
journal = {Information Processing \& Management},
volume = {60},
number = {3},
pages = {103327},
year = {2023},
issn = {0306-4573},
doi = {https://doi.org/10.1016/j.ipm.2023.103327},
url = {https://www.sciencedirect.com/science/article/pii/S030645732300064X},
author = {Weidong Liu and Shuai Li and Yan Cao and Yu Wang},

}

@inproceedings{word2vec2013,
  author       = {Tom{\'{a}}s Mikolov and
                  Kai Chen and
                  Greg Corrado and
                  Jeffrey Dean},
  title        = {Efficient Estimation of Word Representations in Vector Space},
  booktitle    = {1st International Conference on Learning Representations, {ICLR} 2013,
                  Scottsdale, Arizona, USA, May 2-4, 2013, Workshop Track Proceedings},
  year         = {2013},
  
  url          = {http://arxiv.org/abs/1301.3781},
  bibsource    = {dblp computer science bibliography, https://dblp.org}
}

@InProceedings{doc2vec2014,
  title = 	 {Distributed Representations of Sentences and Documents},
  author = 	 {Le, Quoc and Mikolov, Tomas},
  booktitle = 	 {Proceedings of the 31st International Conference on Machine Learning},
  pages = 	 {1188--1196},
  year = 	 {2014},
  editor = 	 {Xing, Eric P. and Jebara, Tony},
  volume = 	 {32},
  number =       {2},
  series = 	 {Proceedings of Machine Learning Research},
  address = 	 {Bejing, China},

  publisher =    {PMLR},
  url = 	 {https://proceedings.mlr.press/v32/le14.html},
  
}

@inproceedings{devlin2019,
    title = "{BERT}: Pre-training of Deep Bidirectional Transformers for Language Understanding",
    author = "Devlin, Jacob  and
      Chang, Ming-Wei  and
      Lee, Kenton  and
      Toutanova, Kristina",
    editor = "Burstein, Jill  and
      Doran, Christy  and
      Solorio, Thamar",
    booktitle = "Proceedings of the 2019 Conference of the North {A}merican Chapter of the Association for Computational Linguistics: Human Language Technologies, Volume 1 (Long and Short Papers)",
    month = jun,
    year = "2019",
    address = "Minneapolis, Minnesota",
    publisher = "Association for Computational Linguistics",
    url = "https://aclanthology.org/N19-1423/",
    doi = "10.18653/v1/N19-1423",
    pages = "4171--4186",

}

@article{BEKAMIRI2024123536,
title = {PatentSBERTa: A deep NLP based hybrid model for patent distance and classification using augmented SBERT},
journal = {Technological Forecasting and Social Change},
volume = {206},
pages = {123536},
year = {2024},
issn = {0040-1625},
doi = {https://doi.org/10.1016/j.techfore.2024.123536},
url = {https://www.sciencedirect.com/science/article/pii/S0040162524003329},
author = {Hamid Bekamiri and Daniel S. Hain and Roman Jurowetzki},
}

@article{STAMATIS2024102282,
title = {A novel re-ranking architecture for patent search},
journal = {World Patent Information},
volume = {78},
pages = {102282},
year = {2024},
issn = {0172-2190},
doi = {https://doi.org/10.1016/j.wpi.2024.102282},
url = {https://www.sciencedirect.com/science/article/pii/S017221902400022X},
author = {Vasileios Stamatis and Michail Salampasis and Konstantinos Diamantaras},
}

@inproceedings{Dai2019,
	title        = {Deeper Text Understanding for IR with Contextual Neural Language Modeling},
	author       = {Dai, Zhuyun and Callan, Jamie},
	year         = 2019,
	month        = jul,
	booktitle    = {Proceedings of the 42nd International ACM SIGIR Conference on Research and Development in Information Retrieval},
	publisher    = {ACM},
	series       = {SIGIR ’19},
	doi          = {10.1145/3331184.3331303},
	url          = {http://dx.doi.org/10.1145/3331184.3331303},
	collection   = {SIGIR ’19}
}

@inproceedings{Piroi2011CLEFIP2R,
  title={CLEF-IP 2011: Retrieval in the Intellectual Property Domain},
  author={Florina Piroi and Mihai Lupu and Allan Hanbury and Veronika Zenz},
  booktitle={Conference and Labs of the Evaluation Forum},
  year={2011},
  url={https://api.semanticscholar.org/CorpusID:12711052}
}

@inbook{Roda2010,
	title        = {CLEF-IP 2009: Retrieval Experiments in the Intellectual Property Domain},
	author       = {Roda, Giovanna and Tait, John and Piroi, Florina and Zenz, Veronika},
	year         = 2010,
	booktitle    = {Multilingual Information Access Evaluation I. Text Retrieval Experiments},
	publisher    = {Springer Berlin Heidelberg},
	pages        = {385--409},
	doi          = {10.1007/978-3-642-15754-747},
	isbn         = 9783642157547,
	issn         = {1611-3349},
	url          = {http://dx.doi.org/10.1007/978-3-642-15754-747}
}

@misc{Marec2021,
  title     = "The {MAREC/IREC} data set",
  author    = "Piroi, Florina",
  publisher = "TU Wien",
  year      =  2021
}

@inproceedings{Sharma2019,
	title        = {BIGPATENT: A Large-Scale Dataset for Abstractive and Coherent Summarization},
	author       = {Sharma, Eva and Li, Chen and Wang, Lu},
	year         = 2019,
	booktitle    = {Proceedings of the 57th Annual Meeting of the Association for Computational Linguistics},
	publisher    = {Association for Computational Linguistics},
	doi          = {10.18653/v1/p19-1212},
	url          = {http://dx.doi.org/10.18653/v1/p19-1212}
}

@article{Fall2003,
	title        = {Automated categorization in the international patent classification},
	author       = {Fall, C. J. and Törcsvári, A. and Benzineb, K. and Karetka, G.},
	year         = 2003,
	month        = apr,
	journal      = {ACM SIGIR Forum},
	publisher    = {Association for Computing Machinery (ACM)},
	volume       = 37,
	number       = 1,
	pages        = {10--25},
	doi          = {10.1145/945546.945547},
	issn         = {0163-5840},
	url          = {http://dx.doi.org/10.1145/945546.945547}
}

@inproceedings{zhang2022contrastive,
    title = "Contrastive Data and Learning for Natural Language Processing",
    author = "Zhang, Rui  and
      Ji, Yangfeng  and
      Zhang, Yue  and
      Passonneau, Rebecca J.",
    editor = "Ballesteros, Miguel  and
      Tsvetkov, Yulia  and
      Alm, Cecilia O.",
    booktitle = "Proceedings of the 2022 Conference of the North American Chapter of the Association for Computational Linguistics: Human Language Technologies: Tutorial Abstracts",
    month = jul,
    year = "2022",
    address = "Seattle, United States",
    publisher = "Association for Computational Linguistics",
    url = "https://aclanthology.org/2022.naacl-tutorials.6/",
    doi = "10.18653/v1/2022.naacl-tutorials.6",
    pages = "39--47",
    
}

\newpage

\appendix

\section*{Appendix — Complete run matrix}
\addcontentsline{toc}{section}{Appendix — Complete run matrix}

\noindent
This appendix reports every configuration in the run matrix
\(\textit{backend} \times \textit{granularity} \times \textit{query view} \times \textit{corpus view} \times \textit{aggregator} \times p\)
with NDCG@100 and Recall@100 on the \textsc{ALL}/\textsc{IN}/\textsc{OUT} subsets.

\medskip
\noindent\textbf{Notation.}
\begin{itemize}
  \item \textbf{Query fields:} \(K=\) Keywords, \(T=\) Title, \(A=\) Abstract, \(C=\) Claims; examples: \(T{+}A\), \(T{+}A{+}C\).
  \item \textbf{Corpus (documents):} \emph{Full Text}, \emph{Title + Abstract}, \emph{Description}, \emph{Title + Abstract + Claims}.
  \item \textbf{Corpus (passages):} \(p\ell\) (passage length \(\ell\)) with aggregator in parentheses:
        \texttt{maxP} (maximum), \texttt{avgP} (mean), \texttt{avg\_topN} (mean of top-\(N = 3\)), \texttt{sumP} (sum).
  \item \textbf{Metrics:} primary = NDCG@100; secondary = Recall@100; each reported for \textsc{ALL}/\textsc{IN}/\textsc{OUT}.
\end{itemize}

\noindent
\emph{Convention.} \emph{Full Text} denotes \emph{Title + Abstract + Claims + Description}.

\tiny
\setlength{\tabcolsep}{2.2pt}
\renewcommand{\arraystretch}{0.9}
\setlength\LTleft{0pt}\setlength\LTright{0pt}

\begin{longtable}{llrrrrrr}
\caption{Full results for \texttt{bm25} — document configurations (ALL/IN/OUT).}
\label{tab:appendix-bm25-document}\\
\toprule
            Query &                    Corpus & NDCG (ALL) & NDCG (IN) & NDCG (OUT) & R@100 (ALL) & R@100 (IN) & R@100 (OUT) \\
\midrule
\endfirsthead
\caption[]{Full results for \texttt{bm25} — document configurations (ALL/IN/OUT).} \\
\toprule
            Query &                    Corpus & NDCG (ALL) & NDCG (IN) & NDCG (OUT) & R@100 (ALL) & R@100 (IN) & R@100 (OUT) \\
\midrule
\endhead
\midrule
\multicolumn{8}{r}{{Continued on next page}} \\
\midrule
\endfoot

\bottomrule
\endlastfoot
        T\,{+}\,A &                 Full Text &     0.2728 &    0.3032 &     0.0525 &      0.3278 &     0.3949 &      0.1368 \\
        T\,{+}\,A &          Title + Abstract &     0.2076 &    0.2335 &     0.0378 &      0.2353 &     0.2894 &      0.0930 \\
        T\,{+}\,A &               Description &     0.2717 &    0.3016 &     0.0524 &      0.3257 &     0.3925 &      0.1362 \\
        T\,{+}\,A & Title + Abstract + Claims &     0.2392 &    0.2680 &     0.0439 &      0.2783 &     0.3432 &      0.1055 \\
T\,{+}\,A\,{+}\,C &                 Full Text &     0.2688 &    0.3006 &     0.0491 &      0.3206 &     0.3870 &      0.1245 \\
T\,{+}\,A\,{+}\,C &          Title + Abstract &     0.1891 &    0.2129 &     0.0306 &      0.2172 &     0.2669 &      0.0833 \\
T\,{+}\,A\,{+}\,C &               Description &     0.2599 &    0.2897 &     0.0465 &      0.3098 &     0.3736 &      0.1180 \\
T\,{+}\,A\,{+}\,C & Title + Abstract + Claims &     0.2451 &    0.2759 &     0.0418 &      0.2820 &     0.3463 &      0.1117 \\
                A &                 Full Text &     0.2636 &    0.2934 &     0.0497 &      0.3159 &     0.3821 &      0.1254 \\
                A &          Title + Abstract &     0.1939 &    0.2182 &     0.0345 &      0.2202 &     0.2739 &      0.0843 \\
                A &               Description &     0.2626 &    0.2920 &     0.0501 &      0.3146 &     0.3797 &      0.1265 \\
                A & Title + Abstract + Claims &     0.2269 &    0.2542 &     0.0413 &      0.2638 &     0.3239 &      0.0993 \\
                K &                 Full Text &     0.2623 &    0.2912 &     0.0522 &      0.3161 &     0.3812 &      0.1286 \\
                K &          Title + Abstract &     0.1905 &    0.2142 &     0.0348 &      0.2196 &     0.2704 &      0.0849 \\
                K &               Description &     0.2613 &    0.2905 &     0.0525 &      0.3153 &     0.3811 &      0.1311 \\
                K & Title + Abstract + Claims &     0.2183 &    0.2458 &     0.0393 &      0.2568 &     0.3177 &      0.0965 \\
\end{longtable}

\tiny
\setlength{\tabcolsep}{2.2pt}
\renewcommand{\arraystretch}{0.9}
\setlength\LTleft{0pt}\setlength\LTright{0pt}

\begin{longtable}{llrrrrrr}
\caption{Full results for \texttt{bm25} — passage configurations (ALL/IN/OUT).}
\label{tab:appendix-bm25-passage}\\
\toprule
            Query &            Corpus & NDCG (ALL) & NDCG (IN) & NDCG (OUT) & R@100 (ALL) & R@100 (IN) & R@100 (OUT) \\
\midrule
\endfirsthead
\caption[]{Full results for \texttt{bm25} — passage configurations (ALL/IN/OUT).} \\
\toprule
            Query &            Corpus & NDCG (ALL) & NDCG (IN) & NDCG (OUT) & R@100 (ALL) & R@100 (IN) & R@100 (OUT) \\
\midrule
\endhead
\midrule
\multicolumn{8}{r}{{Continued on next page}} \\
\midrule
\endfoot

\bottomrule
\endlastfoot
        T\,{+}\,A &        p64 (maxP) &     0.2746 &    0.3079 &     0.0509 &      0.3245 &     0.3912 &      0.1330 \\
        T\,{+}\,A &   p64 (avg\_topN) &     0.0222 &    0.0220 &     0.0084 &      0.0366 &     0.0397 &      0.0235 \\
        T\,{+}\,A &        p64 (avgP) &     0.1861 &    0.2099 &     0.0329 &      0.2291 &     0.2854 &      0.0933 \\
        T\,{+}\,A &        p64 (sumP) &     0.1157 &    0.1268 &     0.0210 &      0.1675 &     0.2029 &      0.0595 \\
        T\,{+}\,A &       p128 (maxP) &     0.2818 &    0.3154 &     0.0533 &      0.3343 &     0.4054 &      0.1391 \\
        T\,{+}\,A &  p128 (avg\_topN) &     0.0222 &    0.0221 &     0.0080 &      0.0366 &     0.0398 &      0.0221 \\
        T\,{+}\,A &       p128 (avgP) &     0.2030 &    0.2304 &     0.0369 &      0.2463 &     0.3071 &      0.1043 \\
        T\,{+}\,A &       p128 (sumP) &     0.1132 &    0.1235 &     0.0214 &      0.1652 &     0.1980 &      0.0609 \\
        T\,{+}\,A &       p256 (maxP) &     0.2863 &    0.3200 &     0.0548 &      0.3390 &     0.4109 &      0.1428 \\
        T\,{+}\,A &  p256 (avg\_topN) &     0.0224 &    0.0221 &     0.0085 &      0.0367 &     0.0396 &      0.0236 \\
        T\,{+}\,A &       p256 (avgP) &     0.2195 &    0.2494 &     0.0403 &      0.2604 &     0.3221 &      0.1023 \\
        T\,{+}\,A &       p256 (sumP) &     0.1072 &    0.1163 &     0.0206 &      0.1587 &     0.1888 &      0.0577 \\
        T\,{+}\,A &       p512 (maxP) &     0.2882 &    0.3220 &     0.0556 &      0.3409 &     0.4128 &      0.1439 \\
        T\,{+}\,A &  p512 (avg\_topN) &     0.0225 &    0.0223 &     0.0085 &      0.0368 &     0.0398 &      0.0235 \\
        T\,{+}\,A &       p512 (avgP) &     0.2348 &    0.2667 &     0.0409 &      0.2769 &     0.3440 &      0.1039 \\
        T\,{+}\,A &       p512 (sumP) &     0.0996 &    0.1076 &     0.0195 &      0.1480 &     0.1759 &      0.0533 \\
        T\,{+}\,A &      p1024 (maxP) &     0.2906 &    0.3241 &     0.0567 &      0.3444 &     0.4169 &      0.1469 \\
        T\,{+}\,A & p1024 (avg\_topN) &     0.0227 &    0.0223 &     0.0085 &      0.0370 &     0.0397 &      0.0229 \\
        T\,{+}\,A &      p1024 (avgP) &     0.2465 &    0.2786 &     0.0450 &      0.2924 &     0.3585 &      0.1174 \\
        T\,{+}\,A &      p1024 (sumP) &     0.0912 &    0.0982 &     0.0176 &      0.1375 &     0.1629 &      0.0480 \\
        T\,{+}\,A &      p2048 (maxP) &     0.2922 &    0.3254 &     0.0579 &      0.3457 &     0.4186 &      0.1489 \\
        T\,{+}\,A & p2048 (avg\_topN) &     0.0227 &    0.0223 &     0.0083 &      0.0370 &     0.0397 &      0.0222 \\
        T\,{+}\,A &      p2048 (avgP) &     0.2532 &    0.2860 &     0.0460 &      0.3023 &     0.3708 &      0.1227 \\
        T\,{+}\,A &      p2048 (sumP) &     0.0830 &    0.0888 &     0.0168 &      0.1260 &     0.1477 &      0.0467 \\
        T\,{+}\,A &      p4096 (maxP) &     0.2929 &    0.3258 &     0.0589 &      0.3468 &     0.4176 &      0.1521 \\
        T\,{+}\,A & p4096 (avg\_topN) &     0.0227 &    0.0224 &     0.0088 &      0.0371 &     0.0403 &      0.0242 \\
        T\,{+}\,A &      p4096 (avgP) &     0.2483 &    0.2790 &     0.0454 &      0.2978 &     0.3631 &      0.1193 \\
        T\,{+}\,A &      p4096 (sumP) &     0.0773 &    0.0825 &     0.0163 &      0.1213 &     0.1432 &      0.0439 \\
        T\,{+}\,A &      p8192 (maxP) &     0.2915 &    0.3244 &     0.0587 &      0.3469 &     0.4184 &      0.1563 \\
        T\,{+}\,A & p8192 (avg\_topN) &     0.0225 &    0.0222 &     0.0085 &      0.0368 &     0.0399 &      0.0231 \\
        T\,{+}\,A &      p8192 (avgP) &     0.2280 &    0.2555 &     0.0418 &      0.2763 &     0.3373 &      0.1143 \\
        T\,{+}\,A &      p8192 (sumP) &     0.0744 &    0.0785 &     0.0164 &      0.1198 &     0.1393 &      0.0456 \\
T\,{+}\,A\,{+}\,C &        p64 (maxP) &     0.2659 &    0.3004 &     0.0451 &      0.3097 &     0.3777 &      0.1238 \\
T\,{+}\,A\,{+}\,C &   p64 (avg\_topN) &     0.0214 &    0.0214 &     0.0074 &      0.0351 &     0.0382 &      0.0207 \\
T\,{+}\,A\,{+}\,C &        p64 (avgP) &     0.1803 &    0.2042 &     0.0307 &      0.2198 &     0.2733 &      0.0870 \\
T\,{+}\,A\,{+}\,C &        p64 (sumP) &     0.1109 &    0.1213 &     0.0206 &      0.1601 &     0.1944 &      0.0588 \\
T\,{+}\,A\,{+}\,C &       p128 (maxP) &     0.2761 &    0.3115 &     0.0477 &      0.3222 &     0.3926 &      0.1313 \\
T\,{+}\,A\,{+}\,C &  p128 (avg\_topN) &     0.0219 &    0.0218 &     0.0080 &      0.0361 &     0.0394 &      0.0225 \\
T\,{+}\,A\,{+}\,C &       p128 (avgP) &     0.2026 &    0.2295 &     0.0353 &      0.2420 &     0.2995 &      0.0952 \\
T\,{+}\,A\,{+}\,C &       p128 (sumP) &     0.1079 &    0.1179 &     0.0197 &      0.1576 &     0.1911 &      0.0559 \\
T\,{+}\,A\,{+}\,C &       p256 (maxP) &     0.2805 &    0.3161 &     0.0484 &      0.3267 &     0.3983 &      0.1323 \\
T\,{+}\,A\,{+}\,C &  p256 (avg\_topN) &     0.0219 &    0.0216 &     0.0083 &      0.0362 &     0.0392 &      0.0228 \\
T\,{+}\,A\,{+}\,C &       p256 (avgP) &     0.2251 &    0.2542 &     0.0394 &      0.2652 &     0.3281 &      0.1040 \\
T\,{+}\,A\,{+}\,C &       p256 (sumP) &     0.1033 &    0.1124 &     0.0191 &      0.1524 &     0.1825 &      0.0529 \\
T\,{+}\,A\,{+}\,C &       p512 (maxP) &     0.2820 &    0.3162 &     0.0496 &      0.3296 &     0.3996 &      0.1333 \\
T\,{+}\,A\,{+}\,C &  p512 (avg\_topN) &     0.0222 &    0.0220 &     0.0082 &      0.0362 &     0.0394 &      0.0215 \\
T\,{+}\,A\,{+}\,C &       p512 (avgP) &     0.2360 &    0.2671 &     0.0409 &      0.2766 &     0.3430 &      0.1052 \\
T\,{+}\,A\,{+}\,C &       p512 (sumP) &     0.0965 &    0.1048 &     0.0180 &      0.1435 &     0.1715 &      0.0499 \\
T\,{+}\,A\,{+}\,C &      p1024 (maxP) &     0.2839 &    0.3179 &     0.0510 &      0.3323 &     0.4040 &      0.1326 \\
T\,{+}\,A\,{+}\,C & p1024 (avg\_topN) &     0.0223 &    0.0220 &     0.0081 &      0.0366 &     0.0397 &      0.0212 \\
T\,{+}\,A\,{+}\,C &      p1024 (avgP) &     0.2465 &    0.2792 &     0.0411 &      0.2888 &     0.3564 &      0.1080 \\
T\,{+}\,A\,{+}\,C &      p1024 (sumP) &     0.0882 &    0.0949 &     0.0182 &      0.1326 &     0.1575 &      0.0525 \\
T\,{+}\,A\,{+}\,C &      p2048 (maxP) &     0.2882 &    0.3222 &     0.0514 &      0.3387 &     0.4105 &      0.1347 \\
T\,{+}\,A\,{+}\,C & p2048 (avg\_topN) &     0.0227 &    0.0222 &     0.0086 &      0.0373 &     0.0402 &      0.0225 \\
T\,{+}\,A\,{+}\,C &      p2048 (avgP) &     0.2564 &    0.2887 &     0.0437 &      0.3024 &     0.3703 &      0.1182 \\
T\,{+}\,A\,{+}\,C &      p2048 (sumP) &     0.0778 &    0.0831 &     0.0175 &      0.1196 &     0.1400 &      0.0513 \\
T\,{+}\,A\,{+}\,C &      p4096 (maxP) &     0.2916 &    0.3265 &     0.0521 &      0.3431 &     0.4158 &      0.1316 \\
T\,{+}\,A\,{+}\,C & p4096 (avg\_topN) &     0.0230 &    0.0224 &     0.0088 &      0.0379 &     0.0405 &      0.0234 \\
T\,{+}\,A\,{+}\,C &      p4096 (avgP) &     0.2557 &    0.2870 &     0.0428 &      0.3033 &     0.3716 &      0.1071 \\
T\,{+}\,A\,{+}\,C &      p4096 (sumP) &     0.0689 &    0.0728 &     0.0162 &      0.1085 &     0.1262 &      0.0433 \\
T\,{+}\,A\,{+}\,C &      p8192 (maxP) &     0.2923 &    0.3275 &     0.0524 &      0.3439 &     0.4175 &      0.1377 \\
T\,{+}\,A\,{+}\,C & p8192 (avg\_topN) &     0.0232 &    0.0225 &     0.0091 &      0.0379 &     0.0403 &      0.0247 \\
T\,{+}\,A\,{+}\,C &      p8192 (avgP) &     0.2421 &    0.2709 &     0.0405 &      0.2883 &     0.3522 &      0.1031 \\
T\,{+}\,A\,{+}\,C &      p8192 (sumP) &     0.0629 &    0.0656 &     0.0147 &      0.1022 &     0.1173 &      0.0390 \\
                A &        p64 (maxP) &     0.2600 &    0.2920 &     0.0475 &      0.3067 &     0.3714 &      0.1214 \\
                A &   p64 (avg\_topN) &     0.0214 &    0.0212 &     0.0078 &      0.0351 &     0.0382 &      0.0222 \\
                A &        p64 (avgP) &     0.1727 &    0.1957 &     0.0303 &      0.2104 &     0.2646 &      0.0869 \\
                A &        p64 (sumP) &     0.1082 &    0.1184 &     0.0206 &      0.1574 &     0.1901 &      0.0599 \\
                A &       p128 (maxP) &     0.2692 &    0.3018 &     0.0508 &      0.3178 &     0.3851 &      0.1321 \\
                A &  p128 (avg\_topN) &     0.0217 &    0.0215 &     0.0079 &      0.0357 &     0.0388 &      0.0226 \\
                A &       p128 (avgP) &     0.1896 &    0.2153 &     0.0330 &      0.2275 &     0.2841 &      0.0927 \\
                A &       p128 (sumP) &     0.1059 &    0.1152 &     0.0205 &      0.1551 &     0.1849 &      0.0582 \\
                A &       p256 (maxP) &     0.2738 &    0.3067 &     0.0524 &      0.3236 &     0.3930 &      0.1336 \\
                A &  p256 (avg\_topN) &     0.0220 &    0.0219 &     0.0079 &      0.0361 &     0.0392 &      0.0225 \\
                A &       p256 (avgP) &     0.2057 &    0.2325 &     0.0390 &      0.2430 &     0.3005 &      0.1001 \\
                A &       p256 (sumP) &     0.1011 &    0.1097 &     0.0195 &      0.1497 &     0.1781 &      0.0542 \\
                A &       p512 (maxP) &     0.2769 &    0.3096 &     0.0534 &      0.3274 &     0.3973 &      0.1364 \\
                A &  p512 (avg\_topN) &     0.0220 &    0.0219 &     0.0078 &      0.0362 &     0.0395 &      0.0220 \\
                A &       p512 (avgP) &     0.2206 &    0.2499 &     0.0397 &      0.2595 &     0.3222 &      0.1003 \\
                A &       p512 (sumP) &     0.0939 &    0.1014 &     0.0190 &      0.1400 &     0.1663 &      0.0533 \\
                A &      p1024 (maxP) &     0.2792 &    0.3115 &     0.0543 &      0.3301 &     0.3996 &      0.1387 \\
                A & p1024 (avg\_topN) &     0.0221 &    0.0218 &     0.0083 &      0.0361 &     0.0389 &      0.0225 \\
                A &      p1024 (avgP) &     0.2318 &    0.2617 &     0.0431 &      0.2739 &     0.3370 &      0.1137 \\
                A &      p1024 (sumP) &     0.0858 &    0.0925 &     0.0170 &      0.1290 &     0.1534 &      0.0462 \\
                A &      p2048 (maxP) &     0.2803 &    0.3124 &     0.0559 &      0.3315 &     0.3999 &      0.1425 \\
                A & p2048 (avg\_topN) &     0.0224 &    0.0220 &     0.0088 &      0.0365 &     0.0398 &      0.0237 \\
                A &      p2048 (avgP) &     0.2397 &    0.2701 &     0.0431 &      0.2856 &     0.3510 &      0.1132 \\
                A &      p2048 (sumP) &     0.0774 &    0.0828 &     0.0156 &      0.1181 &     0.1386 &      0.0415 \\
                A &      p4096 (maxP) &     0.2814 &    0.3134 &     0.0555 &      0.3330 &     0.4009 &      0.1394 \\
                A & p4096 (avg\_topN) &     0.0226 &    0.0223 &     0.0088 &      0.0366 &     0.0401 &      0.0237 \\
                A &      p4096 (avgP) &     0.2352 &    0.2642 &     0.0432 &      0.2806 &     0.3435 &      0.1145 \\
                A &      p4096 (sumP) &     0.0719 &    0.0769 &     0.0148 &      0.1127 &     0.1339 &      0.0378 \\
                A &      p8192 (maxP) &     0.2817 &    0.3140 &     0.0564 &      0.3333 &     0.4011 &      0.1446 \\
                A & p8192 (avg\_topN) &     0.0222 &    0.0219 &     0.0088 &      0.0362 &     0.0395 &      0.0239 \\
                A &      p8192 (avgP) &     0.2164 &    0.2428 &     0.0394 &      0.2605 &     0.3182 &      0.1101 \\
                A &      p8192 (sumP) &     0.0687 &    0.0726 &     0.0153 &      0.1102 &     0.1292 &      0.0419 \\
                K &        p64 (maxP) &     0.2496 &    0.2805 &     0.0441 &      0.2973 &     0.3643 &      0.1148 \\
                K &   p64 (avg\_topN) &     0.0215 &    0.0214 &     0.0075 &      0.0356 &     0.0389 &      0.0210 \\
                K &        p64 (avgP) &     0.1757 &    0.1978 &     0.0322 &      0.2173 &     0.2726 &      0.0872 \\
                K &        p64 (sumP) &     0.1098 &    0.1202 &     0.0212 &      0.1612 &     0.1969 &      0.0570 \\
                K &       p128 (maxP) &     0.2586 &    0.2898 &     0.0473 &      0.3081 &     0.3747 &      0.1231 \\
                K &  p128 (avg\_topN) &     0.0219 &    0.0219 &     0.0076 &      0.0363 &     0.0398 &      0.0221 \\
                K &       p128 (avgP) &     0.1957 &    0.2202 &     0.0360 &      0.2377 &     0.2960 &      0.0971 \\
                K &       p128 (sumP) &     0.1089 &    0.1187 &     0.0210 &      0.1617 &     0.1960 &      0.0564 \\
                K &       p256 (maxP) &     0.2652 &    0.2972 &     0.0494 &      0.3165 &     0.3842 &      0.1274 \\
                K &  p256 (avg\_topN) &     0.0219 &    0.0219 &     0.0075 &      0.0364 &     0.0404 &      0.0218 \\
                K &       p256 (avgP) &     0.2135 &    0.2405 &     0.0397 &      0.2557 &     0.3190 &      0.1034 \\
                K &       p256 (sumP) &     0.1079 &    0.1174 &     0.0208 &      0.1612 &     0.1934 &      0.0551 \\
                K &       p512 (maxP) &     0.2679 &    0.3008 &     0.0504 &      0.3189 &     0.3888 &      0.1273 \\
                K &  p512 (avg\_topN) &     0.0219 &    0.0221 &     0.0072 &      0.0364 &     0.0405 &      0.0206 \\
                K &       p512 (avgP) &     0.2259 &    0.2543 &     0.0419 &      0.2705 &     0.3352 &      0.1070 \\
                K &       p512 (sumP) &     0.1047 &    0.1138 &     0.0209 &      0.1560 &     0.1886 &      0.0564 \\
                K &      p1024 (maxP) &     0.2706 &    0.3032 &     0.0514 &      0.3227 &     0.3928 &      0.1315 \\
                K & p1024 (avg\_topN) &     0.0219 &    0.0220 &     0.0073 &      0.0364 &     0.0403 &      0.0218 \\
                K &      p1024 (avgP) &     0.2359 &    0.2649 &     0.0437 &      0.2835 &     0.3489 &      0.1094 \\
                K &      p1024 (sumP) &     0.1039 &    0.1132 &     0.0205 &      0.1559 &     0.1897 &      0.0556 \\
                K &      p2048 (maxP) &     0.2727 &    0.3049 &     0.0520 &      0.3248 &     0.3951 &      0.1332 \\
                K & p2048 (avg\_topN) &     0.0217 &    0.0216 &     0.0076 &      0.0360 &     0.0390 &      0.0229 \\
                K &      p2048 (avgP) &     0.2428 &    0.2734 &     0.0442 &      0.2903 &     0.3585 &      0.1127 \\
                K &      p2048 (sumP) &     0.1023 &    0.1112 &     0.0199 &      0.1561 &     0.1886 &      0.0544 \\
                K &      p4096 (maxP) &     0.2728 &    0.3053 &     0.0520 &      0.3243 &     0.3947 &      0.1323 \\
                K & p4096 (avg\_topN) &     0.0218 &    0.0216 &     0.0078 &      0.0363 &     0.0392 &      0.0235 \\
                K &      p4096 (avgP) &     0.2434 &    0.2744 &     0.0440 &      0.2907 &     0.3579 &      0.1108 \\
                K &      p4096 (sumP) &     0.1055 &    0.1145 &     0.0212 &      0.1632 &     0.1968 &      0.0596 \\
                K &      p8192 (maxP) &     0.2735 &    0.3058 &     0.0527 &      0.3269 &     0.3974 &      0.1365 \\
                K & p8192 (avg\_topN) &     0.0219 &    0.0218 &     0.0078 &      0.0362 &     0.0401 &      0.0224 \\
                K &      p8192 (avgP) &     0.2339 &    0.2630 &     0.0424 &      0.2813 &     0.3447 &      0.1079 \\
                K &      p8192 (sumP) &     0.1135 &    0.1229 &     0.0230 &      0.1762 &     0.2146 &      0.0653 \\
\end{longtable}

\tiny
\setlength{\tabcolsep}{2.2pt}
\renewcommand{\arraystretch}{0.9}
\setlength\LTleft{0pt}\setlength\LTright{0pt}

\begin{longtable}{llrrrrrr}
\caption{Full results for \texttt{dense} — document configurations (ALL/IN/OUT).}
\label{tab:appendix-dense-document}\\
\toprule
            Query &                    Corpus & NDCG (ALL) & NDCG (IN) & NDCG (OUT) & R@100 (ALL) & R@100 (IN) & R@100 (OUT) \\
\midrule
\endfirsthead
\caption[]{Full results for \texttt{dense} — document configurations (ALL/IN/OUT).} \\
\toprule
            Query &                    Corpus & NDCG (ALL) & NDCG (IN) & NDCG (OUT) & R@100 (ALL) & R@100 (IN) & R@100 (OUT) \\
\midrule
\endhead
\midrule
\multicolumn{8}{r}{{Continued on next page}} \\
\midrule
\endfoot

\bottomrule
\endlastfoot
        T\,{+}\,A &          Title + Abstract &     0.2835 &    0.3207 &     0.0469 &      0.3361 &     0.4144 &      0.1180 \\
        T\,{+}\,A &                  Abstract &     0.2749 &    0.3109 &     0.0448 &      0.3264 &     0.4011 &      0.1124 \\
        T\,{+}\,A & Title + Abstract + Claims &     0.3009 &    0.3411 &     0.0509 &      0.3604 &     0.4423 &      0.1332 \\
T\,{+}\,A\,{+}\,C &          Title + Abstract &     0.2946 &    0.3353 &     0.0448 &      0.3493 &     0.4287 &      0.1221 \\
T\,{+}\,A\,{+}\,C &                  Abstract &     0.2820 &    0.3210 &     0.0432 &      0.3351 &     0.4122 &      0.1153 \\
T\,{+}\,A\,{+}\,C & Title + Abstract + Claims &     0.3055 &    0.3477 &     0.0482 &      0.3627 &     0.4437 &      0.1311 \\
                A &          Title + Abstract &     0.2602 &    0.2948 &     0.0432 &      0.3105 &     0.3831 &      0.1091 \\
                A &                  Abstract &     0.2497 &    0.2831 &     0.0420 &      0.2976 &     0.3682 &      0.1065 \\
                A & Title + Abstract + Claims &     0.2782 &    0.3153 &     0.0462 &      0.3339 &     0.4119 &      0.1209 \\
\end{longtable}

\tiny
\setlength{\tabcolsep}{2.2pt}
\renewcommand{\arraystretch}{0.9}
\setlength\LTleft{0pt}\setlength\LTright{0pt}

\begin{longtable}{llrrrrrr}
\caption{Full results for \texttt{dense} — passage configurations (ALL/IN/OUT).}
\label{tab:appendix-dense-passage}\\
\toprule
            Query &            Corpus & NDCG (ALL) & NDCG (IN) & NDCG (OUT) & R@100 (ALL) & R@100 (IN) & R@100 (OUT) \\
\midrule
\endfirsthead
\caption[]{Full results for \texttt{dense} — passage configurations (ALL/IN/OUT).} \\
\toprule
            Query &            Corpus & NDCG (ALL) & NDCG (IN) & NDCG (OUT) & R@100 (ALL) & R@100 (IN) & R@100 (OUT) \\
\midrule
\endhead
\midrule
\multicolumn{8}{r}{{Continued on next page}} \\
\midrule
\endfoot

\bottomrule
\endlastfoot
        T\,{+}\,A &        p64 (maxP) &     0.2862 &    0.3212 &     0.0508 &      0.3465 &     0.4220 &      0.1292 \\
        T\,{+}\,A &   p64 (avg\_topN) &     0.2893 &    0.3255 &     0.0496 &      0.3519 &     0.4320 &      0.1285 \\
        T\,{+}\,A &        p64 (avgP) &     0.2261 &    0.2564 &     0.0353 &      0.2854 &     0.3551 &      0.0999 \\
        T\,{+}\,A &        p64 (sumP) &     0.0449 &    0.0456 &     0.0137 &      0.0761 &     0.0853 &      0.0408 \\
        T\,{+}\,A &       p128 (maxP) &     0.3009 &    0.3386 &     0.0538 &      0.3639 &     0.4460 &      0.1399 \\
        T\,{+}\,A &  p128 (avg\_topN) &     0.3059 &    0.3447 &     0.0522 &      0.3712 &     0.4556 &      0.1364 \\
        T\,{+}\,A &       p128 (avgP) &     0.2611 &    0.2956 &     0.0423 &      0.3234 &     0.3997 &      0.1164 \\
        T\,{+}\,A &       p128 (sumP) &     0.0336 &    0.0333 &     0.0116 &      0.0572 &     0.0631 &      0.0323 \\
        T\,{+}\,A &       p256 (maxP) &     0.3117 &    0.3502 &     0.0579 &      0.3754 &     0.4574 &      0.1489 \\
        T\,{+}\,A &  p256 (avg\_topN) &     0.3160 &    0.3557 &     0.0562 &      0.3844 &     0.4701 &      0.1439 \\
        T\,{+}\,A &       p256 (avgP) &     0.2838 &    0.3213 &     0.0478 &      0.3488 &     0.4290 &      0.1267 \\
        T\,{+}\,A &       p256 (sumP) &     0.0251 &    0.0244 &     0.0092 &      0.0428 &     0.0467 &      0.0243 \\
        T\,{+}\,A &       p512 (maxP) &     0.3201 &    0.3598 &     0.0576 &      0.3864 &     0.4704 &      0.1414 \\
        T\,{+}\,A &  p512 (avg\_topN) &     0.3234 &    0.3639 &     0.0578 &      0.3917 &     0.4773 &      0.1474 \\
        T\,{+}\,A &       p512 (avgP) &     0.2962 &    0.3350 &     0.0493 &      0.3624 &     0.4456 &      0.1301 \\
        T\,{+}\,A &       p512 (sumP) &     0.0198 &    0.0186 &     0.0081 &      0.0336 &     0.0349 &      0.0214 \\
        T\,{+}\,A &      p1024 (maxP) &     0.3253 &    0.3660 &     0.0592 &      0.3919 &     0.4767 &      0.1538 \\
        T\,{+}\,A & p1024 (avg\_topN) &     0.3279 &    0.3702 &     0.0565 &      0.3975 &     0.4855 &      0.1451 \\
        T\,{+}\,A &      p1024 (avgP) &     0.3058 &    0.3459 &     0.0508 &      0.3732 &     0.4581 &      0.1310 \\
        T\,{+}\,A &      p1024 (sumP) &     0.0149 &    0.0137 &     0.0066 &      0.0257 &     0.0261 &      0.0192 \\
        T\,{+}\,A &      p2048 (maxP) &     0.3274 &    0.3684 &     0.0586 &      0.3964 &     0.4831 &      0.1512 \\
        T\,{+}\,A & p2048 (avg\_topN) &     0.3288 &    0.3710 &     0.0561 &      0.3979 &     0.4873 &      0.1437 \\
        T\,{+}\,A &      p2048 (avgP) &     0.3112 &    0.3525 &     0.0516 &      0.3766 &     0.4649 &      0.1322 \\
        T\,{+}\,A &      p2048 (sumP) &     0.0106 &    0.0095 &     0.0053 &      0.0184 &     0.0176 &      0.0155 \\
        T\,{+}\,A &      p4096 (maxP) &     0.3264 &    0.3676 &     0.0572 &      0.3956 &     0.4828 &      0.1493 \\
        T\,{+}\,A & p4096 (avg\_topN) &     0.3246 &    0.3673 &     0.0542 &      0.3946 &     0.4849 &      0.1405 \\
        T\,{+}\,A &      p4096 (avgP) &     0.3114 &    0.3530 &     0.0521 &      0.3760 &     0.4649 &      0.1319 \\
        T\,{+}\,A &      p4096 (sumP) &     0.0080 &    0.0067 &     0.0044 &      0.0139 &     0.0121 &      0.0133 \\
        T\,{+}\,A &      p8192 (maxP) &     0.3106 &    0.3518 &     0.0510 &      0.3738 &     0.4604 &      0.1296 \\
        T\,{+}\,A & p8192 (avg\_topN) &     0.3135 &    0.3555 &     0.0507 &      0.3781 &     0.4664 &      0.1306 \\
        T\,{+}\,A &      p8192 (avgP) &     0.3012 &    0.3419 &     0.0495 &      0.3608 &     0.4463 &      0.1281 \\
        T\,{+}\,A &      p8192 (sumP) &     0.0066 &    0.0052 &     0.0043 &      0.0113 &     0.0095 &      0.0117 \\
T\,{+}\,A\,{+}\,C &        p64 (maxP) &     0.2910 &    0.3282 &     0.0494 &      0.3513 &     0.4300 &      0.1335 \\
T\,{+}\,A\,{+}\,C &   p64 (avg\_topN) &     0.2954 &    0.3343 &     0.0483 &      0.3590 &     0.4416 &      0.1274 \\
T\,{+}\,A\,{+}\,C &        p64 (avgP) &     0.2361 &    0.2689 &     0.0340 &      0.2936 &     0.3631 &      0.0966 \\
T\,{+}\,A\,{+}\,C &        p64 (sumP) &     0.0451 &    0.0459 &     0.0128 &      0.0758 &     0.0858 &      0.0367 \\
T\,{+}\,A\,{+}\,C &       p128 (maxP) &     0.3057 &    0.3453 &     0.0518 &      0.3692 &     0.4513 &      0.1448 \\
T\,{+}\,A\,{+}\,C &  p128 (avg\_topN) &     0.3122 &    0.3536 &     0.0508 &      0.3780 &     0.4619 &      0.1408 \\
T\,{+}\,A\,{+}\,C &       p128 (avgP) &     0.2719 &    0.3096 &     0.0412 &      0.3349 &     0.4142 &      0.1145 \\
T\,{+}\,A\,{+}\,C &       p128 (sumP) &     0.0342 &    0.0337 &     0.0111 &      0.0577 &     0.0627 &      0.0305 \\
T\,{+}\,A\,{+}\,C &       p256 (maxP) &     0.3207 &    0.3618 &     0.0553 &      0.3870 &     0.4718 &      0.1482 \\
T\,{+}\,A\,{+}\,C &  p256 (avg\_topN) &     0.3244 &    0.3665 &     0.0548 &      0.3926 &     0.4776 &      0.1509 \\
T\,{+}\,A\,{+}\,C &       p256 (avgP) &     0.2968 &    0.3367 &     0.0481 &      0.3633 &     0.4473 &      0.1304 \\
T\,{+}\,A\,{+}\,C &       p256 (sumP) &     0.0249 &    0.0240 &     0.0088 &      0.0424 &     0.0458 &      0.0226 \\
T\,{+}\,A\,{+}\,C &       p512 (maxP) &     0.3293 &    0.3719 &     0.0576 &      0.3963 &     0.4803 &      0.1499 \\
T\,{+}\,A\,{+}\,C &  p512 (avg\_topN) &     0.3329 &    0.3775 &     0.0553 &      0.4009 &     0.4887 &      0.1508 \\
T\,{+}\,A\,{+}\,C &       p512 (avgP) &     0.3112 &    0.3527 &     0.0516 &      0.3787 &     0.4654 &      0.1385 \\
T\,{+}\,A\,{+}\,C &       p512 (sumP) &     0.0194 &    0.0183 &     0.0080 &      0.0331 &     0.0349 &      0.0216 \\
T\,{+}\,A\,{+}\,C &      p1024 (maxP) &     0.3353 &    0.3793 &     0.0577 &      0.4024 &     0.4894 &      0.1515 \\
T\,{+}\,A\,{+}\,C & p1024 (avg\_topN) &     0.3381 &    0.3839 &     0.0555 &      0.4073 &     0.4973 &      0.1524 \\
T\,{+}\,A\,{+}\,C &      p1024 (avgP) &     0.3192 &    0.3622 &     0.0522 &      0.3857 &     0.4739 &      0.1409 \\
T\,{+}\,A\,{+}\,C &      p1024 (sumP) &     0.0141 &    0.0128 &     0.0065 &      0.0241 &     0.0243 &      0.0186 \\
T\,{+}\,A\,{+}\,C &      p2048 (maxP) &     0.3381 &    0.3822 &     0.0590 &      0.4072 &     0.4950 &      0.1552 \\
T\,{+}\,A\,{+}\,C & p2048 (avg\_topN) &     0.3366 &    0.3826 &     0.0541 &      0.4044 &     0.4955 &      0.1480 \\
T\,{+}\,A\,{+}\,C &      p2048 (avgP) &     0.3251 &    0.3691 &     0.0522 &      0.3906 &     0.4795 &      0.1394 \\
T\,{+}\,A\,{+}\,C &      p2048 (sumP) &     0.0101 &    0.0088 &     0.0051 &      0.0175 &     0.0166 &      0.0143 \\
T\,{+}\,A\,{+}\,C &      p4096 (maxP) &     0.3365 &    0.3815 &     0.0564 &      0.4050 &     0.4961 &      0.1500 \\
T\,{+}\,A\,{+}\,C & p4096 (avg\_topN) &     0.3301 &    0.3756 &     0.0523 &      0.3961 &     0.4845 &      0.1445 \\
T\,{+}\,A\,{+}\,C &      p4096 (avgP) &     0.3238 &    0.3683 &     0.0519 &      0.3875 &     0.4772 &      0.1393 \\
T\,{+}\,A\,{+}\,C &      p4096 (sumP) &     0.0079 &    0.0065 &     0.0045 &      0.0139 &     0.0118 &      0.0132 \\
T\,{+}\,A\,{+}\,C &      p8192 (maxP) &     0.3242 &    0.3686 &     0.0528 &      0.3889 &     0.4768 &      0.1435 \\
T\,{+}\,A\,{+}\,C & p8192 (avg\_topN) &     0.3188 &    0.3626 &     0.0502 &      0.3829 &     0.4698 &      0.1371 \\
T\,{+}\,A\,{+}\,C &      p8192 (avgP) &     0.3121 &    0.3551 &     0.0503 &      0.3711 &     0.4577 &      0.1353 \\
T\,{+}\,A\,{+}\,C &      p8192 (sumP) &     0.0068 &    0.0054 &     0.0045 &      0.0116 &     0.0098 &      0.0115 \\
                A &        p64 (maxP) &     0.2570 &    0.2882 &     0.0458 &      0.3154 &     0.3860 &      0.1161 \\
                A &   p64 (avg\_topN) &     0.2627 &    0.2955 &     0.0451 &      0.3225 &     0.3955 &      0.1151 \\
                A &        p64 (avgP) &     0.1980 &    0.2253 &     0.0308 &      0.2538 &     0.3149 &      0.0857 \\
                A &        p64 (sumP) &     0.0419 &    0.0429 &     0.0120 &      0.0710 &     0.0809 &      0.0352 \\
                A &       p128 (maxP) &     0.2713 &    0.3050 &     0.0480 &      0.3296 &     0.4034 &      0.1236 \\
                A &  p128 (avg\_topN) &     0.2789 &    0.3146 &     0.0474 &      0.3405 &     0.4167 &      0.1223 \\
                A &       p128 (avgP) &     0.2324 &    0.2637 &     0.0369 &      0.2902 &     0.3598 &      0.0998 \\
                A &       p128 (sumP) &     0.0319 &    0.0320 &     0.0099 &      0.0541 &     0.0601 &      0.0270 \\
                A &       p256 (maxP) &     0.2833 &    0.3182 &     0.0533 &      0.3441 &     0.4201 &      0.1319 \\
                A &  p256 (avg\_topN) &     0.2899 &    0.3265 &     0.0515 &      0.3534 &     0.4315 &      0.1321 \\
                A &       p256 (avgP) &     0.2569 &    0.2907 &     0.0429 &      0.3177 &     0.3904 &      0.1153 \\
                A &       p256 (sumP) &     0.0240 &    0.0234 &     0.0082 &      0.0410 &     0.0442 &      0.0219 \\
                A &       p512 (maxP) &     0.2928 &    0.3288 &     0.0538 &      0.3569 &     0.4340 &      0.1346 \\
                A &  p512 (avg\_topN) &     0.2985 &    0.3366 &     0.0519 &      0.3640 &     0.4440 &      0.1342 \\
                A &       p512 (avgP) &     0.2706 &    0.3075 &     0.0451 &      0.3333 &     0.4104 &      0.1205 \\
                A &       p512 (sumP) &     0.0191 &    0.0182 &     0.0074 &      0.0324 &     0.0344 &      0.0205 \\
                A &      p1024 (maxP) &     0.2981 &    0.3354 &     0.0539 &      0.3631 &     0.4426 &      0.1382 \\
                A & p1024 (avg\_topN) &     0.3022 &    0.3413 &     0.0524 &      0.3679 &     0.4495 &      0.1375 \\
                A &      p1024 (avgP) &     0.2810 &    0.3188 &     0.0464 &      0.3443 &     0.4232 &      0.1216 \\
                A &      p1024 (sumP) &     0.0145 &    0.0134 &     0.0064 &      0.0249 &     0.0252 &      0.0188 \\
                A &      p2048 (maxP) &     0.3021 &    0.3399 &     0.0558 &      0.3680 &     0.4486 &      0.1449 \\
                A & p2048 (avg\_topN) &     0.3046 &    0.3443 &     0.0518 &      0.3715 &     0.4536 &      0.1349 \\
                A &      p2048 (avgP) &     0.2864 &    0.3254 &     0.0469 &      0.3494 &     0.4315 &      0.1242 \\
                A &      p2048 (sumP) &     0.0102 &    0.0090 &     0.0049 &      0.0175 &     0.0166 &      0.0142 \\
                A &      p4096 (maxP) &     0.3003 &    0.3383 &     0.0539 &      0.3670 &     0.4479 &      0.1434 \\
                A & p4096 (avg\_topN) &     0.3010 &    0.3409 &     0.0509 &      0.3672 &     0.4501 &      0.1345 \\
                A &      p4096 (avgP) &     0.2876 &    0.3273 &     0.0468 &      0.3507 &     0.4340 &      0.1228 \\
                A &      p4096 (sumP) &     0.0076 &    0.0065 &     0.0044 &      0.0133 &     0.0119 &      0.0131 \\
                A &      p8192 (maxP) &     0.2850 &    0.3222 &     0.0486 &      0.3472 &     0.4273 &      0.1230 \\
                A & p8192 (avg\_topN) &     0.2899 &    0.3295 &     0.0478 &      0.3525 &     0.4358 &      0.1225 \\
                A &      p8192 (avgP) &     0.2780 &    0.3166 &     0.0449 &      0.3373 &     0.4177 &      0.1171 \\
                A &      p8192 (sumP) &     0.0061 &    0.0049 &     0.0039 &      0.0105 &     0.0091 &      0.0103 \\
\end{longtable}

\end{document}